\documentclass[ijds,nonblindrev]{informs-ijds}

\OneAndAHalfSpacedXI

\usepackage{latexsym,amssymb,amsmath,amsbsy,amsopn,amstext,amsxtra,color,bm,calc,ifpdf} 
\usepackage{array}
\usepackage{float}
\usepackage{enumerate}
\usepackage{fancyhdr}
\usepackage{listings}
\usepackage{multicol}
\usepackage{multirow}
\usepackage{makeidx}
\usepackage{makecell}
\usepackage{xcolor}
\usepackage{tabularx,booktabs}
\graphicspath{{figures/}}
\DeclareGraphicsExtensions{.pdf,.jpeg,.png}	
\usepackage[caption=false,font=footnotesize]{subfig}
\usepackage{url}
\usepackage{algorithm}
\usepackage{algorithmicx}
\usepackage{algpseudocode}
\usepackage{appendix}

\newcommand {\Eqref}[1]{Eq. (\ref{#1})}
\newcommand {\Figref}[1]{Fig. \ref{#1}}
\newcommand {\Tabref}[1]{TABLE \ref{#1}}
\newcommand {\Romnum}[1]{\uppercase\expandafter{\romannumeral #1}}

\usepackage{hyperref}  
\usepackage{natbib}
 \bibpunct[, ]{(}{)}{,}{a}{}{,}%
 \def\bibfont{\small}%
 \def\BIBand{and}%

\TheoremsNumberedThrough     
\ECRepeatTheorems

\EquationsNumberedThrough    

\MANUSCRIPTNO{IJDS}

\begin{document}


\RUNAUTHOR{Wang et al.} 

\RUNTITLE{Non-stationary and Sparsely-correlated MGP}

\TITLE{Non-stationary and Sparsely-correlated Multi-output Gaussian Process with Spike-and-Slab Prior\footnote{The code capsule has been submitted to Code Ocean with provisional DOI: 10.24433/CO.4010696.v1. }}

\ARTICLEAUTHORS{%
\AUTHOR{Xinming Wang}
\AFF{Department of Industrial Engineering and Management, Peking University, China. \EMAIL{wang-xm20@stu.pku.edu.cn}} 
\AUTHOR{Yongxiang Li}
\AFF{Department of Industrial Engineering and Management, Shanghai Jiaotong University, China}
\AUTHOR{Xiaowei Yue}
\AFF{Department of Industrial Engineering, Tsinghua University, China.}
\AUTHOR{Jianguo Wu}
\AFF{Department of Industrial Engineering and Management, Peking University, China. \EMAIL{j.wu@pku.edu.cn}} 
} 

\ABSTRACT{%
Multi-output Gaussian process (MGP) is commonly used as a transfer learning method to leverage information among multiple outputs. A key advantage of MGP is providing uncertainty quantification for prediction, which is highly important for subsequent decision-making tasks. However, traditional MGP may not be sufficiently flexible to handle multivariate data with dynamic characteristics, particularly when dealing with complex temporal correlations. Additionally, since some outputs may lack correlation, transferring information among them may lead to negative transfer. To address these issues, this study proposes a non-stationary MGP model that can capture both the dynamic and sparse correlation among outputs. Specifically, the covariance functions of MGP are constructed using convolutions of time-varying kernel functions. Then a dynamic spike-and-slab prior is placed on correlation parameters to automatically decide which sources are informative to the target output in the training process. An expectation-maximization (EM) algorithm is proposed for efficient model fitting. Both numerical studies and a real case demonstrate its efficacy in capturing dynamic and sparse correlation structure and mitigating negative transfer for high-dimensional time-series data. {\color{black} Finally, a mountain-car reinforcement learning case highlights its potential application in decision making problems.}
}%



\maketitle

%


\section{Introduction}
Gaussian process (GP) provides an elegant and flexible Bayesian non-parametric framework for modeling nonlinear mappings \citep{GP2006}. Characterized solely by mean and covariance functions, it is capable of capturing complex input-output relationships, as well as measuring prediction uncertainty which is critical for decision-making. As a result, GP has been widely applied in various fields, such as 
Bayesian optimization \citep{BayesianOptimization2018}, experiment design  \citep{Surrogates2020}, and product quality monitoring \citep{QualityMonitoring2016}. 
However, standard GP is designed for only one single output, which limits its use in multi-output or multi-task scenarios arising in various fields, such as Bayesian optimization \citep{MGPBO2023}, traffic network \citep{MGPtraffic2019}, and computer simulation emulator \citep{MGPemulator2013}. Consequently, multi-output Gaussian process (MGP) has been gaining increasing attention from researchers and has emerged as an important member in the vast family of transfer learning \citep{TransferSurvey2009} and multi-task learning \citep{Multi-taskSurvey2021} methods. 

Stationary GP and MGP models are commonly used with covariance function depending only on the distance among data points. However, this invariance to input space translation makes them unsuitable for non-stationary environments, where the data characteristics vary across the input domain \citep{GP2006}. 
This phenomenon is quite common in time-series data. For instance, in the energy field, the mean of power consumption in a household is different in every season \citep{Non-stationaryGP2023}. 
In clinical studies, sepsis is very likely to cause changes in the cross-correlation among vital signs in the early onset \citep{VitalSign2017}. 
In kinesiology, the cooperation patterns of human joints vary across different gestures, e.g., both hands move jointly in a `shoot' action but separately in a `throw' action \citep{DynamicSubspace2022}. 
In such cases, non-stationary models, that allow all or a subset of parameters to vary are generally more appropriate. \textcolor{black}{Modeling and capturing such a structural change are important in subsequent decision making tasks, such as identifying the risk of disease and taking a medical care \citep{Non-stationaryGP2023}.}

Mainly two kinds of methods have been proposed to capture the dynamic characteristics of the non-stationary data. 
The first category assumes that the parameters are the same within local regions but different across regions. For example, a Bayesian tree-based GP \citep{TreeGP2008} uses a tree structure to partition the input space of computer simulation emulator. Another method, called jump GP, cuts the input space into several segments to model piece-wise continuous functions \citep{JumpGP2022}. This model is optimized using the expectation-minimization (EM) algorithm or variational inference. Besides, a clustering-based GP \citep{ClusterGP2017} partitions the spatial data into groups by calculating a cluster dissimilarity and constructs stationary GPs for each group of data. A space-partitioning based GP is further extended to the active learning area to accelerate the design of experiments of heterogeneous systems \citep{ActivePartitionGP2023}. However, these methods are not suitable for data with gradually-changing characteristics. 
To address this issue, the methods in the second category abandon the locally-stationary assumption. They allow all or some parameters to be input/time-dependent, and model those parameters by additional GPs.  For instance, non-stationary GP introduced in \citep{Non-stationaryGP2016} and \citep{Non-stationaryGP2023} applies GP priors on the amplitude and length-scale parameters of a square-exponential covariance function. 
In addition, based on these single-output non-stationary GPs, researchers have explored MGP models for multivariate data with dynamic characteristics. For example, a non-stationary MGP is established to model the varying correlation between vital signals, where a GP prior is imposed on the time-dependent correlation matrix \citep{Non-stationaryMGP2021}. 
However, in this state-of-the-art MGP model, the GP prior has no shrinkage effect, which does not encourage a sparse estimation of cross-correlation among multiple outputs.

Pursuing sparse estimation of cross-correlation is rooted in the negative information transfer, which is another critical challenge when using MGP. Transfer learning is very promising for leveraging information to data-insufficient domain, called target domain, from other correlated domains, called source domain. However, not all the data from the source domain are necessarily correlated with the target domain. If knowledge is transferred from uncorrelated domains, it may reduce the performance of target learning, known as negative transfer \citep{TransferSurvey2009, PositiveTransfer2018}. For example, the motion signal of a specific human joint may only be correlated with a subset of the other joints. In order to recover the joint's motion information by borrowing information from others, it is necessary to detect which joints share similar moving trajectories with the target joint. Therefore, it is crucial that researchers or engineers can make the best choice on using which sources to transfer information. 

Negative transfer exists widely in transfer learning, often stemming from the excessive inclusion of source data. To handle this issue, one straightforward approach is to measure the relatedness of each source to the target and choose the most related one for information transfer. For example, the method proposed in \citep{TransferForecasting2021} takes Jensen-Shannon (JS) divergence as a criterion and selects the source with the least divergence for knowledge transfer. However, such a method only takes the pairwise transferability into account and ignores the global structure between the target and the sources. And this choice is made independently on the specific model before training, which is far from an optimal decision. An alternative approach is the regularized MGP \citep{RegularizedMGP2022}, which jointly models all outputs and selects informative sources during the training process. However, all these approaches assume that the source-target cross-correlation is fixed in the time space, and thus cannot model the dynamic and sparse structure among multiple outputs.

To this end, we propose a non-stationary MGP model to capture the varying characteristics of data and mitigate the negative transfer simultaneously. Specifically, we focus on modeling the dynamic and sparse correlations between the sources and the target. 
In the proposed framework, we first construct a convolution-process-based MGP for transfer learning, whose covariance function parameters are allowed to vary in time space. We then apply a spike-and-slab prior to the parameters that are related to the sparse correlation between the sources and the target. The slab part mainly accounts for smoothly-changing or constant correlation parameters, while the spike part is responsible for shrinking some parameters to zero, thereby removing the corresponding uninformative sources.
To the best of our knowledge, this is the first research on MGP that simultaneously handles dynamic relationship and negative transfer.
Our contributions can be summarized as follows:
\begin{enumerate}
\item
A novel non-stationary MGP model is established using the convolution of latent processes and time-dependent kernel functions, which is suitable for modeling multiple outputs with varying characteristics.
\item
\textcolor{black}{A dynamic spike-and-slab prior is applied to capture the temporal and sparse correlations among outputs, deciding from which sources to transfer information to the target.}
\item
The mixture weight of the spike and slab priors is automatically adjusted during the training process using an EM-based optimization algorithm, which can effectively prevent placing shrinkage effects on non-zero elements. 
\end{enumerate}

The rest of this paper is organized as follows. In Section \ref{sec: preliminaries}, we revisit the related literature and the static MGP model. Section \ref{sec: model development} presents the proposed non-stationary MGP and an efficient EM algorithm for model training. {\color{black} In Section \ref{sec: numerical study}, we evaluate the effectiveness of our model on simulated data. In Section \ref{sec: case study}, we perform one time-series analysis case on human gesture data \citep{GestureData2012} and one control policy optimization case on the mountain-car problem \citep{MountainCar1990}.} In Section \ref{sec: conclusion}, we conclude the paper with a discussion.

\section{Preliminaries}
\label{sec: preliminaries}
In this section,  we first review researches that are related to our work. We then introduce the static MGP based on the convolution process, which has been widely applied in various areas due to its flexibility \citep{DependentGP2004, RULMGP2018, OnlineMGP}. 

\subsection{Related work}
To deal with non-stationary data, a natural extension of GPs is to release the restriction that the parameters of the covariance functions are invariant throughout the input space. Most of the existing approaches either encourage the parameters to be constant in a local area and construct a piece-wise model (the first category, e.g., \citep{TreeGP2008, ClusterGP2017, ActivePartitionGP2023, JumpGP2022}), %
or allow them to vary at each point and model them using other GPs (the second category, e.g.,  \citep{Non-stationaryGP2003, Non-stationaryGP2016, Non-stationaryGP2023}). The methods in the second category have a similar structure to that of a two-layer Deep Gaussian Process (Deep GP), where the input is first transformed by the first GP layer into a latent input, and then fed into the second GP layer to obtain the output \citep{DeepGP2013, DeepGP2022, BOdeepGP2022}. However, the parameters of Deep GP are stationary, which differs from the second category where the covariance parameters are dynamic.

Non-stationary GPs mainly focus on modeling the dynamic mean, smoothness, and amplitude parameters. With regards to MGP, dynamic correlation is another key characteristic that needs to be considered.  In classical Linear model of Coregionalization (LMC), each output is a linear combination of several latent Gaussian processes, and the covariance matrix is modeled by a Kronecker product of a correlation matrix and a single GP’s covariance matrix \citep{LMC1992}.The existing non-stationary MGPs are mainly extensions of the classical LMC model. 
For example, the approach in \citep{VaryingCoregionalization2004} allows the correlation matrix to vary with inputs to model the dynamic relationship among outputs. 
In \citep{Non-stationaryMGP2021}, a non-stationary MGP combines the time-varying correlation (across outputs) and smoothness (within each output) together. However, as extensions of the traditional LMC, these methods also suffer from the limitation that all outputs possess the same covariance structure. More flexible MGPs are proposed by constructing each output through the convolution process and modeling them with individual parameters \citep{DependentGP2004}. However, these approaches are for stationary data. Furthermore, all existing approaches fail to capture a sparse correlation structure in a non-stationary environment.

{\color{black} Spatial-temporal modeling of non-stationary data is closely related to our work. In comparison with normal time-series modeling, spatial-temporal analysis requires to model the spatial correlation to enhance the prediction accuracy. A large number of spatial-temporal models have been investigated, such as spatial-temporal auto-regressive integrated moving average method (ST-ARIMA) \citep{ARIMA2003}, spatial-temporal k-nearest neighbors (ST-KNN) \citep{STKNN2016}, spatial-temporal random fields \citep{STRandomFields2012, STRandomFields2022}, and spatial-temporal deep neural networks \citep{STDeep2020, TransformersSurvey2023}. Based on the aforementioned methods, a number of recent works try to extend them to handle non-stationary spatial-temporal data. One popular and efficient solution is utilizing some change detection algorithm to partition the time-series into several stationary periods, and then applying the stationary model for each period, e.g., a ST-KNN with a wrapped K-means partition algorithm \citep{NonStaSTKNN2021}, an auto-regressive model coupled with a block-fused-Lasso change-point detection algorithm \citep{NonStaAR2022}. Besides partitioning the time-series into stationary parts, the method proposed by \citep{NonStaAugment2017} maps the non-stationary space-time process into a high-dimensional stationary process through augmenting new dimensions. Another type of non-stationary spatial-temporal model is Bayesian random fields with non-stationary space-time kernels \citep{NonStaSTGP2012, NonStaFourier2018, NonStaRandomFields2020, NonStaRandomFields2023}, whose hyper-parameters change over time or location. Deep learning methods are also explored on non-stationary data recently, such as non-stationary recurrent neural networks \citep{NonStaRNN2018, NonStaRNN2020}, long short-term memory networks \citep{NonStaLSTM2019}, and transformer-based networks \citep{NonStaTransformers2022, TransformersSurvey2023}. In contrast to the spatial-temporal model, MGP does not impose a restriction that the source outputs must be sampled during the same period as the target outputs. Furthermore, it does not depend on spatial distance to establish correlations among outputs. As a result, the MGP model is capable of accommodating a wider range of scenarios. }

It is important to mention that we use a \emph{dynamic} spike-and-slab prior in our model. The classical spike-and-slab prior is a Bayesian selection approach \citep{SS1993} that has been used for feature selection in (generalized) linear models, additive models, and Gaussian processes \citep{SS2005, FunctionSelection2012, SSGP2022}. With this prior, smaller parameters tend to be more influenced by the spike prior to reach zero, while the larger ones are mainly dominated by the slab part and bear little shrinkage. However, the classical spike-and-slab prior cannot account for the modeling of dynamic and sparse correlation parameters in our model. Therefore, we propose to extend this prior to a dynamic version. Although current works have explored the dynamic variable selection for the varying-coefficient linear models \citep{SparseTVP2014, SparseTVP2021, DynamicSS2021}, no work utilizes a dynamic spike-and-slab prior to model the dynamic and sparse correlation among outputs in a non-stationary MGP.


\subsection{Static MGP based on convolution process}
\label{sec: static MGP}
Consider a set of $m$ outputs ${f}_i : \mathcal{X} \mapsto \mathbb{R},\ i=1,...,m$, where $\mathcal{X}$ is a $d$-dimensional input domain applied to all outputs. Suppose that the observation $y_i$ is accompanied with independent and identically distributed (i.i.d.) noise $\epsilon_i \sim \mathcal{N}(0, \sigma_i^2)$, i.e.,
$$y_i (\bm{x}) = f_i(\bm{x}) + \epsilon_i.$$
where $\bm{x} \in \mathbb{R}^d$ is the input.
Denote the $n_i$ observed data for the $i$th output as $\mathcal{D}_i=\left\{\bm{X}_i, \bm{y}_i\right\}$, where $\bm{X}_i=(\bm{x}_{i,1},...,\bm{x}_{i,n_i})^T$ and $\bm{y}_i=(y_{i,1},...,y_{i,n_i})^T$ are the collections of input points and associated observations respectively. Let the total number of observations be represented by $N = \sum_i n_i$ for $m$ outputs. Denote the data of all outputs as $\mathcal{D}=\left\{\bm{X}, \bm{y}\right\}$, where $\bm{X}=\left(\bm{X}_1^T,...,\bm{X}_m^T\right)^T \in \mathbb{R}^{N \times d} $ and $\bm{y}=\left(\bm{y}_1^T,...,\bm{y}_m^T \right)^T \in \mathbb{R}^{N }$. 
In an MGP model, the observation vector $\bm{y}$ follows a joint Gaussian distribution:
\begin{align}
 \bm{y}| \bm{X}  \sim \mathcal{N}
\begin{pmatrix}
	 \bm{0}, \bm{K}
\end{pmatrix},
\label{eq: MGP}
\end{align}
where $\bm{K}=\bm{K}(\bm{X},\bm{X}) \in \mathbb {R}^{N \times N}$ is a block-partitioned covariance matrix. 
The $(i,i^{\prime})$-th block of $\bm{K}$, $\bm{K}_{i,i^{\prime}}={\rm cov}_{i,i^{\prime}}^f(\bm{X}_i,\bm{X}_{i^{\prime}}) + \tau_{i, i^{\prime}} \sigma_i^2 \bm{I} \in \mathbb{R}^{n_i \times n_{i^{\prime}}}$ represents the covariance matrix between the output $i$ and output $i^{\prime}$ ($\tau_{i, i^{\prime}}$ equals to $1$ if $i = i^{\prime}$, and $0$ otherwise). The function ${\rm cov}_{i,i^{\prime}}^f(\bm{x}, \bm{x}^{\prime})$ measures the covariance between $f_i(\bm{x})$ and $f_{i^{\prime}}(\bm{x}^{\prime})$. In the covariance matrix $\bm{K}$, the cross-covariance block $\bm{K}_{i,i^{\prime}}(i \neq i^{\prime})$ is the most important part to realize information transfer, as it models the correlation between different outputs.

As the convolution of a GP and a smoothing kernel is still a GP, we can construct each output $f_i$ through convolving a group of shared latent processes $\{z_j(\bm{x})\}_{j=1}^h$ and kernel functions $\{g_{ji}(\bm{x})\}_{j=1}^h$  in the following way \citep{DependentGP2004, EfficientConvolvedMGP2011}:
\begin{align}
f_i(\bm{x}) = \sum_{j = 1}^h \alpha_{ji}g_{ji}(\bm{x})\ast z_j(\bm{x}) =  \sum_{j = 1}^h \alpha_{ji}\int_{-\infty}^{\infty} g_{ji}(\bm{x}-\bm{u}) z_j(\bm{u}) d \bm{u}
\label{eq: MGP convolution structure}
\end{align}
where $\ast$ represents convolution operation, $\alpha_{ji}$ is the amplitude parameter, and $h$ is the number of shared latent processes. Usually, $\{z_j(\bm{x})\}_{j=1}^h$ are independent white Gaussian noise processes with ${\rm cov}(z(\bm{x}), z(\bm{x}^{\prime})) = \delta(\bm{x}-\bm{x}^{\prime})$, where $\delta(\cdot)$ is the Dirac delta function. Thus, the covariance function can be derived as:
\begin{align}
{\rm cov}_{i,i^{\prime}}^f (\bm{x}, \bm{x}^{\prime})  = {\rm cov}[f_i(\bm{x}),  f_{i^{\prime}}(\bm{x}^{\prime})]  
&= \sum_{j=1}^h{\rm cov}\{ \alpha_{ji} g_{ji}(\bm{x})\ast z_j (\bm{x}), \alpha_{ji^{\prime}} g_{ji^{\prime}}(\bm{x}^{\prime})\ast z_j (\bm{x}^{\prime})\} \notag\\
&=\sum_{j=1}^h  \alpha_{ji}\alpha_{ji^{\prime}} \int_{-\infty}^{\infty} g_{ji}(\bm{u})g_{ji^{\prime}}(\bm{u}-\bm{v})d \bm{u},
\label{eq:cov in convolution process}
\end{align}
where $\bm{v}=\bm{x}-\bm{x}^{\prime}$. 
It such a way, the covariance between $f_i(\bm{x})$ and $f_{i^{\prime}}(\bm{x}^{\prime})$ is dependent on their difference $\bm{x}-\bm{x}^{\prime}$, the amplitude parameters, and the hyperparameters in kernels $g_{ji}$ and $g_{ji^{\prime}}$. Compared with the classical LMC model $f_i(\bm{x}) = \sum_{j = 1}^h \alpha_{ji} q_j(\bm{x})$,
where $q_j(\bm{x}) $ is a latent GP with covariance $k_j(\bm{x}, \bm{x}^{\prime})$. The convolution-process-based MGP is more flexible than LMC, as it does not restrict all outputs to having the same auto-covariance pattern.

At a new point $\bm{x}_*$, the posterior distribution of $y_i(\bm{x}_*)$ given data $\{\bm{X},\bm{y}\}$ is:
\begin{align}
y_i(\bm{x}_*)| \bm{X},\bm{y} \sim \mathcal{N}\left( \mu(\bm{x}_*), {\Sigma}(\bm{x}_*) \right),
\label{eq: predictive distribution}
\end{align}
where the predictive mean $\mu(\bm{x}_*)$ and variance $\Sigma(\bm{x}_*)$ can be expressed as:
\begin{align}
\label{eq:mean prediction}
\mu(\bm{x}_*)&=\bm{K}_*^T \bm{K}^{-1} \bm{y}, \\
\label{eq:variance prediction}
\Sigma(\bm{x}_*)&={\rm cov}_{ii}^f(\bm{x}_*, \bm{x}_*)  + \sigma_i^2 - \bm{K} _*^T \bm{K}^{-1}\bm{K}_*,
\end{align}
where $\bm{K}_*^T = [{\rm cov}_{i1}^f(\bm{x}_*, \bm{X}_1)^T, ..., {\rm cov}_{im}^f(\bm{x}_*, \bm{X}_m)^T] $ is the covariance between the new point $x_*$ and all observed data. From the posterior distribution, we can find that the covariance function plays a crucial role in prediction. For instance, the predicted mean is the linear combination of output data, where the weight is decided by the covariance matrix. However, in the static MGP, the covariance between two data points depends solely on their distance and does not change dynamically. Additionally, some outputs may be uncorrelated with others, therefore the estimated covariance matrix should possess a sparse structure to avoid negative transfer between the uncorrelated outputs. In the following section, we will propose a novel non-stationary MGP to simultaneously address both problems.

\section{Model Development}
\label{sec: model development}
We propose a non-stationary MGP model for transfer learning that can capture sparse source-target correlation in a dynamic environment. Specifically, we assume that the correlations between the target and each source vary over time. Besides, some sources may not be related to the target during certain time periods. 
Under such a circumstance, a spike-and-slab prior is utilized to model the varying and sparse correlation structure. 

\subsection{The proposed model.}
\begin{figure*}[!t]
\centering
\caption{The graphical structure of non-stationary MGP. Latent processes and kernel functions are with a gray background, while the parameters are with a white background. The parameters' priors are shown in rectangles.}
\includegraphics[width=6.0in]{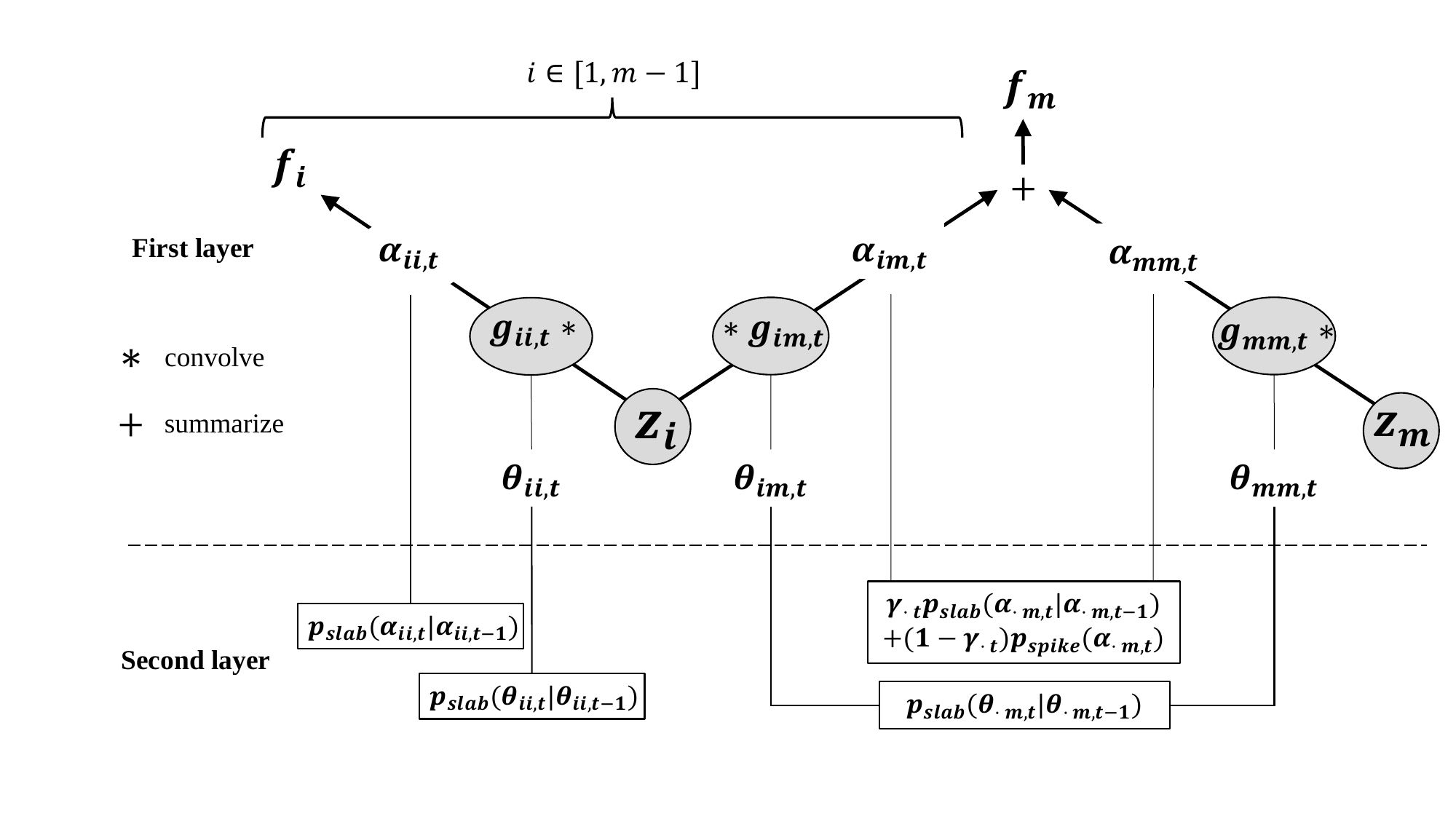}
\label{fig: graphical structure}
\end{figure*}

The structure of our hierarchical model is illustrated in \Figref{fig: graphical structure}. The first layer constructs outputs through the convolution of time-dependent kernel functions and latent white Gaussian noise processes, and the second layer consists of priors on function parameters designed to encourage desired properties, such as smoothness and sparsity.

\subsubsection{Dynamic MGP}
In this subsection, we introduce the major part of the proposed non-stationary MGP, which corresponds to the first layer in \Figref{fig: graphical structure}.
To simplify the notation, we slightly abuse the notation used in the previous section. Specifically, we take the first $m-1$ outputs ${f}_i : \mathcal{X} \mapsto \mathbb{R},\ i=1,...,m-1$ as the sources, and the last one ${f}_m : \mathcal{X} \mapsto \mathbb{R}$ as the target. We still assume that the observation $y_i$ is accompanied with the i.i.d. measurement noise $\epsilon_i \sim N(0, \sigma_i^2)$. Let $\mathcal{I}=\{1,2,...,m\}$ be the index set of all outputs, and  $\mathcal{I}^S=\mathcal{I}/\{m\}$ contain the indices of all sources. For simplicity yet without loss of generality, we assume the source and target data are sampled at the time $t \in \{1, 2, ... ,n\}$, i.e., $\bm{X}_i = (\bm{x}_{i, 1}, ..., \bm{x}_{i, t}, ..., \bm{x}_{i, n})^T$ and $\bm{y}_{i} = (y_{i,1},...,y_{i,t},..,y_{i,n})^T$. In Appendix A, we will show a more general case where each output is observed only at a subset of time stamps. 

Based on the above assumption, our dynamic MGP model is formulated as:
\begin{align}
y_{i} (\bm{x}_t) & = f_{i}(\bm{x}_t) + \epsilon_i =\alpha_{ii, t} g_{ii, t}(\bm{x}_t)\ast z_{i}(\bm{x}_t) + \epsilon_i , i \in \mathcal{I}^S \notag\\
y_{m} (\bm{x}_t) & = f_{m}(\bm{x}_t) + \epsilon_m =\sum_{j \in \mathcal{I}}\alpha_{jm, t}g_{jm, t}(\bm{x}_t) \ast z_{j}(\bm{x}_t) + \epsilon_m  
\label{eq: dynamic MGP}
\end{align}
where $\alpha_{ii, t}$ and $\alpha_{jm,t}$ are time-varying amplitude parameters, $g_{ii, t}(\bm{x})$ and $g_{jm, t}(\bm{x})$ are time-varying kernel functions, and $\{z_{i}(\bm{x})\}_{i=1}^{m}$ are latent white Gaussian noise processes independent of each other. 
This model is highly flexible as various types of kernel functions can be utilized. We choose to employ a Gaussian kernel which is widely used due to its flexibility \citep{DependentGP2004, EfficientConvolvedMGP2011}. The kernel is given by 
\begin{align}
g_{ij, t}(\bm{x})= (2\pi)^{-\frac{d}{4}}|\bm{\theta}_{ij, t}|^{-\frac{1}{4}}\exp \left(-\frac{1}{2}\bm{x}^T\bm{\theta}_{ij, t}^{-1}\bm{x} \right),
\label{eq: Gaussian kernel}
\end{align}
where $\bm{\theta}_{ij, t}$ is a diagonal matrix representing the length-scale for each input feature. {\color{black} More importantly, such a Gaussian kernel can yield closed-from covariance functions through the convolution operation in \Eqref{eq:cov in convolution process} \citep{Non-stationaryGP2003, NonStaRandomFields2020}. }

This flexible model allows each source to have its own kernel $g_{ii}$, thereby allowing for heterogeneity among the sources. In order to transfer knowledge from the sources to the target, the target is connected to $\{z_i\}_{i=1}^{m-1}$ though the kernel function $g_{im, t}$.  Regarding the parameters, $\{\alpha_{ii, t}, \bm{\theta}_{ii, t}\}_{i=1}^m$ are responsible for the non-stationary behavior within each output, while $\{\alpha_{im, t}, \bm{\theta}_{im, t}\}_{i=1}^{m-1}$ capture the dynamic correlation between target and sources. More specifically,  the amplitude parameter $\alpha_{im, t}$ controls the knowledge transfer. For example, if $\alpha_{im, t}=0$, then  $f_i$ will not transfer information to $f_m$ at time $t$.

As the latent processes are independent of each other, the covariance matrix among sources is zero-valued. Therefore, the covariance matrix can be re-partitioned as:
\begin{align}
\bm{K}=
\renewcommand{\arraystretch}{1.0}
\setlength{\arraycolsep}{1.0pt}
\begin{pmatrix}
\begin{array}{ccc|c}
\bm{K}_{1,1}  & \cdots & \bm{0} & \bm{K}_{1,m} \\
\vdots  & \ddots & \vdots & \vdots \\
\bm{0}  & \cdots & \bm{K}_{m-1,m-1} & \bm{K}_{m-1,m} \\ 
\hline
\bm{K}_{1,m}^T   & \cdots & \bm{K}_{m-1,m}^T & \bm{K}_{m,m}
\end{array} 
\end{pmatrix}
=
\renewcommand{\arraystretch}{1.0}
\begin{pmatrix} \begin{array}{c|c}
\bm{K}_{(ss)} & \bm{K}_{(sm)} \\ \hline \bm{K}_{(sm)}^T & \bm{K}_{mm}
\end{array} \end{pmatrix},
\label{eq: covariance matrix}
\end{align}
where the $(i,i^{\prime})$-th block $\bm{K}_{i,i^{\prime}}={\rm cov}_{i,i^{\prime}}^f(\bm{X}_i,\bm{X}_{i^{\prime}}) + \tau_{i, i^{\prime}} \sigma_i^2 \bm{I}$, the block-diagonal matrix $\bm{K}_{(ss)}$ represents the covariance of source outputs, and $\bm{K}_{(sm)}$ represents the cross-covariance between the sources and the target. 
Based on \Eqref{eq:cov in convolution process},
we can obtain covariance functions for the proposed non-stationary MGP model, as shown below:
\begin{subequations}\label{eq: dynamic cov collection}
\begin{align}
&{\rm cov}_{ii}^f(\bm{x}_t, \bm{x}_{t^{\prime}}) \notag\\
=& \alpha_{ii, t}\alpha_{ii, t^{\prime}} 
{\frac{|\bm{\theta}_{ii, t}|^{\frac{1}{4}} |\bm{\theta}_{ii, t^{\prime}}|^{\frac{1}{4}}} {|\bm{\theta}_{ii, t} + \bm{\theta}_{ii, t^{\prime}}|^{\frac{1}{2}}}} 
\exp \left[ {-\frac{1}{2}} {(\bm{x}_t - \bm{x}_{t^{\prime}})^T} {(\bm{\theta}_{ii, t} + \bm{\theta}_{ii, t^{\prime}})^{-1}} {(\bm{x}_t - \bm{x}_{t^{\prime}})}\right], 
\label{eq: dynamic source auto-cov}\\
&{\rm cov}_{im}^f(\bm{x}_t, \bm{x}_{t^{\prime}}) \notag\\
=& \alpha_{ii, t}\alpha_{im, t^{\prime}}
{\frac{|\bm{\theta}_{ii, t}|^{\frac{1}{4}} |\bm{\theta}_{im, t^{\prime}}|^{\frac{1}{4}}} {|\bm{\theta}_{ii, t} + \bm{\theta}_{im, t^{\prime}}|^{\frac{1}{2}}}} 
\exp \left[ {-\frac{1}{2}} {(\bm{x}_t - \bm{x}_{t^{\prime}})^T} {(\bm{\theta}_{ii, t} + \bm{\theta}_{im, t^{\prime}})^{-1}} {(\bm{x}_t - \bm{x}_{t^{\prime}})}\right], 
\label{eq: dynamic cross-cov}\\
&{\rm cov}_{mm}^f(\bm{x}_t, \bm{x}_{t^{\prime}}) \notag\\
=&\sum_{j \in \mathcal{I}} \alpha_{jm, t}\alpha_{jm, t^{\prime}} 
{\frac{|\bm{\theta}_{jm, t}|^{\frac{1}{4}} |\bm{\theta}_{jm, t^{\prime}}|^{\frac{1}{4}}} {|\bm{\theta}_{jm, t} + \bm{\theta}_{jm, t^{\prime}}|^{\frac{1}{2}}}} 
\exp \left[ {-\frac{1}{2}} {(\bm{x}_t - \bm{x}_{t^{\prime}})^T} {(\bm{\theta}_{jm, t} + \bm{\theta}_{jm, t^{\prime}})^{-1}} {(\bm{x}_t - \bm{x}_{t^{\prime}})}\right], 
\label{eq: dynamic target auto-cov}
\end{align}
\end{subequations}
where $i \in \mathcal{I}^S$.
Equations (\ref{eq: dynamic source auto-cov}- \ref{eq: dynamic target auto-cov}) represent the covariance within the sources, between the sources and the target, and within the target, respectively. To ensure the positivity of those hyper-parameters, we utilize a soft-plus transformation for them \citep{Non-stationaryGP2016}:
$\alpha_{ij,t} = \log[1 + \exp(\tilde{\alpha}_{ij,t})],  \bm{\theta}_{ij,t} = \log[1 + \exp(\tilde{\bm{\theta}}_{ij,t})],$
where $\tilde{\alpha}_{ij,t}, \tilde{\bm{\theta}}_{ij,t}$ are underlying parameters to estimate, whose range is $[-\infty, \infty]$. 
{\color{black} The proposed covariance functions can be viewed as an extension of the non-stationary kernels \citep{Non-stationaryGP2003} from the single-output to the multi-output case. Specifically, the auto-covariance of each source in \Eqref{eq: dynamic source auto-cov} is the same as the covariance for a single-output non-stationary GP.} From the cross-covariance between each source and the target, we can clearly see that the amplitude parameter $\alpha_{im, t}$ controls whether the cross-correlation is zero or not. The validity of the proposed covariance functions is outlined in Proposition \ref{proposition: positive definiteness}:
\begin{proposition}
The proposed non-stationary MGP covariance matrix in \Eqref{eq: covariance matrix} is positive-definite, i.e., $\forall \bm{y} \neq \bm{0}$,
$$\bm{y}^T \bm{K} \bm{y} > 0.$$\vspace {-2.0em}
\label{proposition: positive definiteness}
\end{proposition}\noindent 
The proof is provided in Appendix B. 

Based on \Eqref{eq: covariance matrix}, the  joint distribution of all sources and the target is expressed as:
\begin{align}
\renewcommand{\arraystretch}{1.0}
\setlength{\arraycolsep}{1.0pt}
\begin{pmatrix} \begin{array}{c}
\bm{y}_{(s)}\\ \bm{y}_{m}
\end{array}  \end{pmatrix}
\Big| \bm{X}
\sim \mathcal{N}
\left(
\begin{bmatrix} \begin{array} {c}
\bm{0} \\ \bm{0}
\end{array} \end{bmatrix},
\begin{bmatrix} \begin{array}{cc}
\bm{K}_{(ss)} & \bm{K}_{(sm)} \\ \bm{K}_{(sm)}^T & \bm{K}_{ mm}
\end{array} \end{bmatrix}
\right),
\label{eq: joint distribution}
\end{align}
where $\bm{y}_{(s)}$ is the collection of all source data. For notational convenience, we partition the parameters in a similar way, 
$
\bm{\alpha}_{(s)} = \{\bm{\alpha}_{(s), t}\}_{t=1}^n,
\bm{\theta}_{(s)} = \{\bm{\theta}_{(s), t}\}_{t=1}^n,
\bm{\sigma}_{(s)} = \{\sigma_i\}_{i=1}^{m-1},
\bm{\alpha}_m = \{\bm{\alpha}_{m, t}\}_{t=1}^n ,
\bm{\theta}_m = \{\bm{\theta}_{m, t}\}_{t=1}^n,
$
where $\bm{\alpha}_{(s), t}=\{\alpha_{ii, t}\}_{i=1}^{m-1}$, $\bm{\theta}_{(s), t}=\{\theta_{ii, t}\}_{i=1}^{m-1}$, $\bm{\alpha}_{m, t}=\{\alpha_{im, t}\}_{i=1}^{m}$,  and $\bm{\theta}_{m, t} = \{\theta_{im, t}\}_{i=1}^{m}$. Furthermore, we denote the collection of all parameters as $\bm{\Phi} = \{ \bm{\Phi}_{(s)}, \bm{\Phi}_{m}\}$, where $\bm{\Phi}_{(s)} = \{\bm{\alpha}_{(s)}, \bm{\theta}_{(s)}, \bm{\sigma}_{(s)}\}$ and $ \bm{\Phi}_{m} = \{\bm{\alpha}_{m}, \bm{\theta}_{m}, {\sigma}_{m}\}$. 

In the proposed model, the most important and challenging task is to estimate those time-varying parameters. If no restriction is applied, model training may suffer from a serious over-fitting problem. To address this issue, Gaussian processes are typically employed to model the kernel parameters, e.g., 
$\log(\alpha_{t}) \sim \mathcal{GP}(0, k_{\alpha}(t, t^{\prime})), \log(\bm{\theta}_{t}) \sim \mathcal{GP}(0, k_{\bm{\theta}}(t, t^{\prime}))$, where $k_{\alpha},  k_{\bm{\theta}}$ are covariance functions for the amplitude and length-scale parameters respectively. Although this technique can force the parameters to vary smoothly and reduce over-fitting, it cannot model a sparse correlation between the sources and the target. Consequently, this approach cannot avoid the negative transfer caused by unrelated sources. 

\subsubsection{Dynamic spike-and-slab prior.}
\textcolor{black}{The classical spike-and-slab prior \citep{SS1993, SSGP2022} only handles the shrinkage of one single parameter and cannot model the smooth-varying parameters.
To take both the dynamic and sparse property of correlation into account,} we propose a dynamic spike-and-slab prior placing on $\bm{\alpha}_{m}$: 
\begin{align}
\alpha_{im, t}|\gamma_{i, t}, \alpha_{im, t-1} & \sim(1-\gamma_{i, t}) {p}_{spike}(\alpha_{im, t})+\gamma_{i, t} {p}_{slab}(\alpha_{im, t}| \alpha_{im, t-1}) \notag\\
{\gamma}_{i, t}|{\eta} & \sim  Bern(\eta), 
\label{eq: dynamic spike-and-slab prior}
\end{align}
where $\gamma_{i, t}\in\{0,1\}$ is a binary sparse indicator for $\alpha_{im, t}$ following a Bernoulli distribution, $ {p}_{spike}(\alpha_{im, t})$ is a zero-mean spike prior pushing parameter to zero, ${p}_{slab}(\alpha_{t, im}| \alpha_{t-1, im})$ is a slab prior connecting $\alpha_{im, t-1}$ and $\alpha_{im, t}$, and $\eta$ is a prior weight between the spike and slab. If there is no prior information regarding the weight, we can set $\eta$ to $0.5$. The spike-and-slab prior is shown in the second layer of the graphical structure in \Figref{fig: graphical structure}. As for all the other parameters, we do not force them to possess sparsity, so only the slab prior is placed on them to control the smoothness, i.e.,
\begin{align}
\alpha_{ii, t}| \alpha_{ii, t-1} & \sim {p}_{slab}(\alpha_{ii, t}| \alpha_{ii, t-1}),  \notag\\
\bm{\theta}_{ii, t}| \bm{\theta}_{ii, t-1} & \sim {p}_{slab}(\bm{\theta}_{ii, t}| \bm{\theta}_{ii, t-1}),  \notag\\
\bm{\theta}_{im, t}| \bm{\theta}_{im, t-1} & \sim {p}_{slab}(\bm{\theta}_{im, t}| \bm{\theta}_{im, t-1}).
\label{eq: dynamic slab prior}
\end{align}
Note that $\bm{\theta}_{ij, t}$ is a diagonal matrix, so that the slab prior is placed on its $d$ diagonal elements independently, i.e., ${p}_{slab}(\bm{\theta}_{ii, t}| \bm{\theta}_{ii, t-1}) = \prod_{l=1}^d {p}_{slab}(\{\bm{\theta}_{ii, t}\}_{l}| \{\bm{\theta}_{ii, t-1}\}_{l}) $.
By using the conditional distributions as priors, we can control the change of amplitude or smoothness from the previous time step to the current one, e.g., from $\alpha_{ii, t-1}$ to $\alpha_{ii, t}$.
Compared with the dynamic spike-and-slab prior used in linear models \citep{DynamicSS2021}, our method does not constrain the slab prior to be a stable autoregressive process. Besides, we use a simpler but more flexible prior for $\gamma_{i,t}$, while the work in  \citep{DynamicSS2021} uses a prior conditional on the coefficients of the linear model.

The spike prior is responsible for shrinking the parameters to zero and cutting down the information transfer channel to the target. Common choices for this prior include point mass distribution, Laplace distribution, and Gaussian distribution with a small variance. Considering the shrinkage performance and optimization convenience, we choose the Laplace distribution as the spike prior, i.e.,
\begin{align}
{p}_{spike}(\alpha_{im, t}) = \frac{1}{2\nu_0} \exp \left(-\frac{| \alpha_{im, t} |}{\nu_0}\right),  
\label{eq: spike prior}
\end{align}
where $\nu_0$ is the length-scale for Laplace distribution. If we maximize the log-likelihood function to optimize the model, this prior will become a $L_1$ norm penalty and have the ability to shrink parameters.
The slab prior encourages the smoothness of parameter change. In this work, we consider two types of slab priors. The first one is a hard slab prior, 
\begin{align}
{p}_{slab}^{hard}(\alpha_{im, t}| \alpha_{im, t-1}) = \frac{1}{2\nu_1} \exp \left(-\frac{| \alpha_{im, t} -  \alpha_{im, t-1}|}{\nu_1}\right),
\label{eq: hard slab}
\end{align}
which encourages $\alpha_{im, t}$ to remain constant in a continuous period, approximating a piecewise model. In the second one, the parameters are allowed to change smoothly,
\begin{align}
{p}_{slab}^{soft}(\alpha_{im, t}| \alpha_{im, t-1}) =  \frac{1}{\sqrt{2 \pi \nu_1}} \exp \left(-\frac{( \alpha_{im, t} - \rho \alpha_{im, t-1})^2}{2\nu_1}\right),
\end{align}
where $\nu_1$ is variance of Gaussian distribution, and $\rho < 1$ is an autoregressive coefficient.  A similar smoothing approach can also be found in \citep{OnlineMGP}. These two slab priors make the current parameter value exactly or roughly concentrated around the previous value. Typically, we set $\nu_0$ to be much smaller than $\sqrt{\nu_1}$ in the soft slab prior (or $\nu_1$ in the hard slab prior) to make the two priors more separable and to put more penalty on sparsity. Besides, the values of $\bm{\alpha}_{im}$ at multiple time steps before $t$ can be included in the soft slab prior, e.g., ${p}_{slab}^{soft}(\alpha_{im, t}| \alpha_{im, t-1}, \alpha_{im, t-2}, ...)$. We choose the simplest form ${p}_{slab}^{soft}(\alpha_{im, t}| \alpha_{im, t-1})$ due to its wide application and robust performance in practice. 


At time $t$, $\eta$ can be interpreted as the prior probability that $\alpha_{im, t}$ belongs to the slab process. It influences the strength of shrinkage effect that $\alpha_{im, t}$ bears. Ideally, for non-zero $\alpha_{im, t}$, the posterior mean of $\gamma_{i,t}$ should be close to one so that $\alpha_{im, t}$ is barely impacted by the spike prior. In the optimization algorithm developed in the next subsection, we will show that the estimated mean of $\gamma_{i,t}$ is automatically adjusted based on the estimated $\alpha_{im, t}$ to avoid shrinking non-zero elements. This makes our method superior to traditional Lasso methods where the sparse penalty weights are identical for zero and non-zero parameters.

Finally, based on the above discussion, the whole hierarchical model of the proposed non-stationary MGP can be expressed as follows:
\begin{align}
&\bm{y}_{(s)}, \bm{y}_m | \bm{\Phi}_{(s)}, \bm{\Phi}_m \sim \mathcal{N} (\bm{0}, \bm{K}) \notag\\  
& \alpha_{ii, t}| \alpha_{ii, t-1}  \sim {p}_{slab}(\alpha_{ii, t}| \alpha_{ii, t-1}),  i\in \mathcal{I}^{S},\notag\\
&\bm{\theta}_{ii, t}| \bm{\theta}_{ii, t-1} \sim {p}_{slab}(\bm{\theta}_{ii, t}| \bm{\theta}_{ii, t-1}),  i\in \mathcal{I}^{S}, \notag\\
& \alpha_{im, t}|\gamma_{i, t}, \alpha_{im, t-1} \sim (1-\gamma_{i, t}) {p}_{spike}(\alpha_{im, t})  +\gamma_{i, t} {p}_{slab}(\alpha_{im, t}| \alpha_{im, t-1}), i\in \mathcal{I}, \notag\\
& {\gamma}_{i, t}|{\eta}  \sim  Bern(\eta), i\in \mathcal{I}, \notag\\
&\bm{\theta}_{im, t}| \bm{\theta}_{im, t-1} \sim {p}_{slab}(\bm{\theta}_{im, t}| \bm{\theta}_{im, t-1}), i\in \mathcal{I}.
\label{eq: dynamic generative model}
\end{align}

\subsection{Expectation-maximization-based optimization algorithm}
\label{sec: EM}
\textcolor{black}{The widely-used algorithm for a Bayesian model is Markov Chain Monte Carlo (MCMC) sampling, but it is computationally-inefficient for the proposed non-stationary model with considerable time-varying parameters. Therefore, we develop an efficient EM algorithm.} Instead of directly maximizing the posterior $p(\bm{\Phi}|\bm{y})=p({\bm{\Phi}}_{(s)}, \bm{\Phi}_m | \bm{y}_{(s)}, \bm{y}_m)$, we proceed iteratively in terms of the complete log posterior $\log p({\bm{\Phi}}, \bm{\gamma} | \bm{y})$, where the binary parameters $\bm{\gamma}$ are treated as ``missing data''. Since this function is not observable, in the Expectation-step (E-step), we calculate its conditional expectation given the observed data and the current estimated parameters. Then in the Maximization-step (M-step), we maximize the expected complete log-posterior with respect to $\bm{\Phi}$.
More precisely, the E-step and M-step at the $(k+1)$th iteration can be expressed as:
\begin{align}
&{\rm E-step: } \ Q\left(\bm{\Phi} | {\bm{\Phi}}^{(k)}\right) = E_{\bm{\gamma} | {\bm{\Phi}}^{(k)}, \bm{y}} \left\{ \log p({\bm{\Phi}}, \bm{\gamma} | \bm{y})\right\}, \notag\\
&{\rm M-step: } \ {\bm{\Phi}}^{(k+1)} = \underset{\bm{\Phi}}{\rm argmax} \left\{ Q\left(\bm{\Phi} | {\bm{\Phi}}^{(k)}\right) \right\}
\end{align}
where $E_{\bm{\gamma} | {\bm{\Phi}}^{(k)}, \bm{y}} (\cdot)$ is the conditional expectation on posterior of $\bm{\gamma}$, and ${\bm{\Phi}}^{(k)}$ is the optimized parameters at the $k$th iteration. For simplicity, we use $E_{\bm{\gamma}} (\cdot)$ to denote $E_{\bm{\gamma} | {\bm{\Phi}}^{(k)}, \bm{y}} (\cdot)$. 

Based on Bayes' Theorem and the property of multivariate normal distribution, the expectation of $\log p({\bm{\Phi}}, \bm{\gamma} | \bm{y})$ can be as (derivation details can be found in Appendix C):
\begin{align}
E_{\bm{\gamma}} \left\{ \log p({\bm{\Phi}}, \bm{\gamma} | \bm{y}) \right\} 
=& -\frac{1}{2} \left\{ \bm{y}_{(s)}^T \bm{K}_{(ss)}^{-1} \bm{y}_{(s)} + \log|\bm{K}_{(ss)}| + (\bm{y}_{m}-\bm{\mu})^T \bm{\Sigma}^{-1} (\bm{y}_{m}-\bm{\mu}) + \log|\bm{\Sigma}| \right\} \notag\\
& + \sum_{i=1}^{m-1} \sum_{t=2}^n \Big[  \log {p}_{slab}(\bm{\theta}_{ii, t}| \bm{\theta}_{ii, t-1}) + \log {p}_{slab}({\alpha}_{ii, t}| {\alpha}_{ii, t-1})   \Big] \notag\\
& + \sum_{i=1}^m \sum_{t=2}^n \Big[ \log {p}_{slab}(\bm{\theta}_{im, t}| \bm{\theta}_{im, t-1}) + (1-E_{\bm{\gamma}}{\gamma}_{i, t}) \log {p}_{spike}(\alpha_{im, t})  \notag\\
&\quad \qquad \qquad + E_{\bm{\gamma}}{\gamma}_{i, t} \log {p}_{slab}(\alpha_{im, t}| \alpha_{im, t-1}) \Big] + const., 
\label{eq: EM objective}
\end{align}
where $\bm{\mu}=\bm{K}_{(sm)}^T \bm{K}_{(ss)}^{-1}\bm{{y}}_{(s)}$ is the conditional mean of target given the sources and $\bm{\Sigma}=\bm{K}_{mm}-\bm{K}_{(sm)}^T \bm{K}_{(ss)}^{-1}\bm{K}_{(sm)}$ is the conditional covariance. 

In the E-step, since $\bm{\gamma}$ is only dependent on $\bm{\Phi}_{m}$, the posterior of $\gamma_{i, t}$ is calculated as:
\begin{align}
&p(\gamma_{i, t}| {\bm{\Phi}}^{(k)}_{m}) = \frac{p({\alpha}_{im, t}^{(k)} | \gamma_{i, t}) p(\gamma_{i, t})}{p({\alpha}_{im, t}^{(k)})} \propto p({\alpha}_{im, t}^{(k)} | \gamma_{i, t}) p(\gamma_{i, t}) \notag\\
&=\left[ (1-\gamma_{i, t}) {p}_{spike}({\alpha}_{im, t}^{(k)}) +\gamma_{i, t} {p}_{slab}({\alpha}_{im, t}^{(k)}| {\alpha}_{im, t-1}^{(k)}) \right]  \cdot {\eta}^{\gamma_{i, t}} {(1-\eta)}^{(1-\gamma_{i, t})}.
\end{align}
Then the conditional expectation of $\gamma_{i, t}$ can be updated as:
\begin{align}
E_{\bm{\gamma}} {\gamma}_{i, t} = \frac{\eta {p}_{slab}({\alpha}_{im, t}^{(k)}| {\alpha}_{im, t-1}^{(k)})}
{(1-\eta) {p}_{spike}({\alpha}_{im, t}^{(k)}) +\eta {p}_{slab}({\alpha}_{im, t}^{(k)}| {\alpha}_{im, t-1}^{(k)})},
\label{eq: gamma update}
\end{align}

The posterior mean $E_{\bm{\gamma}} {\gamma}_{i, t} $ can be interpreted as the posterior probability of classifying the current parameter ${\alpha}_{im, t}$ into a slab process as opposed to a spike process, based on the past value $\alpha_{im, t-1}$. For example, we set $\eta$ to 0.5 as a non-informative prior, and take a small $\nu_0$ (e.g., 0.01) for the spike prior and a large $\nu_1$ (e.g., 0.1) for the soft slab prior. Based on \Eqref{eq: gamma update}, supposing that $(\alpha_{im, t}-\alpha_{im, t-1})^2$ is small and $|\alpha_{im, t}|$ is large, the expectation $E_{\bm{\gamma}}{\gamma}_{i, t}$ will tend towards one, indicating that $\alpha_{im, t}$ is more likely from the slab prior. 
On the other hand, if $|\alpha_{im, t}|$ is small, $E_{\bm{\gamma}}{\gamma}_{i, t}$ is close to zero, enforcing strong shrinkage on $\alpha_{im, t}$. 

In the M-step, we can optimize the objective function in \Eqref{eq: EM objective} with various gradient ascent methods, such as (stochastic) ADAM, L-BFGS, etc. \citep{ADAM2014, BFGS1997}. 
This objective function is actually a standard Gaussian process log-likelihood with additional regularization terms. The regularization terms penalize the difference between parameters at successive time points and shrink the amplitude parameters to facilitate source selection. The weights of the regularization terms are modulated by the expectation of $\bm{\gamma}$. Ideally, for non-zero $\alpha_{im, t}$, $E_{\gamma}{\gamma}_{i,t}$ will equal to one, so no shrinkage effect will be placed on it. In other words, the strength of sparse penalty is automatically adjusted through \Eqref{eq: gamma update}, and this adjustment has explicit statistical interpretability. 

In our case studies, we find the algorithm converges rapidly, e.g., achieving convergence after only five iterations. 
The whole algorithm is summarized in Algorithm \ref{algorithm: summary}, where an ADAM method \citep{ADAM2014} is utilized in the M-step.
Note that the large parameter space of $\bm{\Phi}$ poses a challenge in identifying the modes of the full posterior. To speed up the convergence, we propose to initialize the source parameter $\bm{\Phi}_{(s)}$ by maximizing the sum of sources’ marginal log-likelihood and source parameter prior:
\begin{align}
\underset{\bm{\Phi}_{(s)}} {\rm max}\ &\log p(\bm{y}_{(s)}|  \bm{\Phi}_{(s)}) + \log p ( \bm{\Phi}_{(s)} ) 
\label{eq: source initialization}
\end{align}
For target parameters, we find simple random initialization can achieve satisfactory performance in experiments.

\subsection{Computational Challenge}
\label{subsec: computational challenge}
{\color{black}
There are three main computational challenges that we need to address. The first one is the calculation of integration for a convolution kernel. To avoid an intractable integration for the covariance, we utilize the Gaussian kernel in \Eqref{eq: Gaussian kernel} and derive closed-form covariance functions in Equations (\ref{eq: dynamic source auto-cov}- \ref{eq: dynamic target auto-cov}).

The second challenge is calculating the inverse of covariance matrix, which is a critical issue for all GPs. The computational complexity of Algorithm \ref{algorithm: summary} is approximately $O(\sum_{i=1}^{m} n_i^3)$, where $n_i$ is the number of data points for each output. In comparison, the computational cost for traditional MGP is $O((\sum_{i=1}^{m} n_i)^3)$ with the same size of data and non-separable covariance.  The main computational load of our model is in calculating the inverse of $\bm{K}_{(ss)}$ and $\bm{\Sigma}$ in the source marginal distribution  $\log p(\bm{y}_{(s)}|  \bm{\Phi}_{(s)} )$ and the target conditional distribution $\log  p(\bm{y}_m | \bm{\Phi}_m, \bm{y}_{(s)}, {\bm{\Phi}}_{(s)})$ respectively in \Eqref{eq: EM objective}. Since all the latent processes are independent of each other, $\bm{K}_{(ss)}$ is a block-diagonal matrix and the complexity is reduced from $O(\sum_{i=1}^{m-1} n_i^3)$ to $O(\sum_{i=1}^{m-1} n_i^3)$. The calculation of the inverse of $\bm{\Sigma}$ is $O(n_m^3)$. Therefore, the overall computational complexity is reduced to $O(\sum_{i=1}^{m} n_i^3)$.

Finally, it is a hard task to estimate a considerable number of time-varying parameters. Therefore, we develop the EM-based algorithm to fit the model rather than using a sampling method. Based on the results of a non-stationary linear model \citep{DynamicSS2021}, the MCMC and EM algorithm lead to very close prediction errors, but the running time of MCMC is about ten times longer than that of the EM.
}
\begin{algorithm}
\caption{The optimization algorithm for the non-stationary MGP model}
\label{algorithm: summary}
\begin{algorithmic}[1]
\Require Data $\{ \bm{X}_i, \bm{y}_{i}\}_{i=1}^{m}$, $\nu_0$, $\nu_1$, $\rho$, $\eta$.
\State Set starting value: ${\bm{\Phi}}_{(s)}^{(-1)}, {\bm{\Phi}}_{m}^{(0)}$.
\State Initialization: obtain ${\bm{\Phi}}_{(s)}^{(0)}$ through \Eqref{eq: source initialization}. 
\For{$k_{out}$ iterations}
	\State \textbf{E-step} given ${\bm{\Phi}}^{(k)}$:  Update $\{E_{\bm{\gamma}}{\gamma}_{i,t}\}_{i=1,t=2}^{m, n}$ through \Eqref{eq: gamma update}.
	\State \textbf{M-step} given $E_{\bm{\gamma}}{\gamma}_{i,t}$:
	\For{$k_{in}$ epochs}
		\State Calculate $\bm{K}_{(ss)}$, $\bm{K}_{(sm)}$, and $\bm{K}_{mm}$  according to \Eqref{eq: dynamic cov collection}.
		\State Calculate the value and gradient of objective function in \Eqref{eq: EM objective}.
		\State Obtain ${\bm{\Phi}}^{(k+1)}$ using the ADAM method.
	\EndFor
\EndFor
\end{algorithmic}
\end{algorithm}

\subsection{Tuning Parameter Selection}
{\color{black} The tuning parameters for our model is $\nu_0$, $\nu_1$ (for the hard slab prior), and $\nu_2$ (for the soft slab prior). Here, since the key of our method is selecting the most informative sources, we propose to maximize the following criterion:
\begin{align}
    B(\bm{\nu}) = N \log p(\bm{y}|\bm{X}) - \log(N) c_{\bm{\nu}}(\bm{\alpha}_m),
\end{align}
where $N$ is the number of data, $\log p(\bm{y}|\bm{X})$ is the log-likelihood for both the source and the target outputs, and $c_{\bm{\nu}}(\bm{\alpha}_m)$ is the number of nonzero elements in $\bm{\alpha}_m$ given $\bm{\nu}$. Similar criterion is proposed in \citep{DynamicSubspace2021}. Note that $\bm{\nu} = \{ \nu_0, \nu_1\}$ for the hard slab prior and  $\bm{\nu} = \{ \nu_0, \nu_2\}$ for the soft slab prior. This criterion is similar to the Bayesian Information Criterion (BIC). The first term tends to select a more complex model with a larger likelihood for all outputs, while the second term prefers simpler models where less sources chosen to transfer information to the source. To reduce the computation, we first determine the ratios $r_1 = \nu_1/\nu_0$ and $r_2 = \nu_2/\nu_0$ to make the spike and slab priors separable \citep{DynamicSS2021}. Then we design a two-dimension search grid for $(\nu_1, \nu_1/r_1)$ with the hard slab or $(\nu_2, \nu_2/r_1)$ with the soft slab. The optimal value of $\bm{\nu}$ is searched over the two-dimensional grid. For example, we set $r_1\in\{5, 10\}$ and $(\nu_1, \nu_0) \in \{(1/5, 1/25), (1/5, 1/50), (1/10, 1/50), (1/10, 1/100), (1/15, 1/75), (1/15, 1/150)\}$ for a hard slab prior.}

\subsection{Model Prediction}
Since the EM algorithm only estimates the value of parameters at the observed time points, given a new input $\bm{x}_{{t}^*}$ of interest, we first need to estimate the target parameter $\bm{\alpha}_{m, t^*}$ and  $\bm{\theta}_{m, t^*}$ at the new time point $t^*$, then derive the predictive distribution of $y_m (\bm{x}_{t^{*}})$. 

\subsubsection{Forecasting.}
In the forecasting task, $t^* > n$. The estimated value of $\bm{\theta}_{im, t^*}$ is,
\begin{align*}
&\bm{\theta}_{im, t^*} = \left\{
\begin{matrix}
 \bm{\theta}_{im, n},  & {\rm for \ hard \ slab \ prior}, \\
\rho^{t^*-n} \bm{\theta}_{im, n}, & {\rm for\ soft\ slab\ prior},
\end{matrix}
\right.
\end{align*}
which is actually the mode of $p_{slab}(\bm{\theta}_{im, t^*} | \bm{\theta}_{im, n})$. As for ${\alpha}_{im, t^*}$, if $E_{\bm{\gamma}}\gamma_{i, n}\geq 0.5$, we consider it from the slab process and estimate it using the same method as $\bm{\theta}_{im, t^*}$. Otherwise, it is classified to a spike process and shrunk to zero \citep{EMVS2014}. 

\subsubsection{Recovery.}
In the recovery task, the target data are unobserved at some specific time points and we aim to recover the missing data, i.e., $1<t^*<n$. Define $t_{be}$ and $t_{af}$ to be the nearest observed time points before and after ${t}^*$ respectively. Denote the nearest observation time to $t^*$ as $t_{near}$, i.e., 
$$t_{near} = \underset{t\in\{t_{be}, t_{af}\}}{\rm argmin} |t-{t}^*| .$$
As the parameters before and after $t_*$ are already optimized by the EM algorithm, the estimation of $\bm{\theta}_{im, t^*}$ becomes:
\begin{align*}
&\bm{\theta}_{im, t^*} = \left\{
\begin{matrix}
 \bm{\theta}_{im, t_{near}},  & {\rm for \ hard \ slab \ prior}, \\
{\rm LSE} (\bm{\theta}_{im, t_{be}}, \bm{\theta}_{im, t_{af}}), & {\rm for\ soft\ slab\ prior},
\end{matrix}
\right.
\end{align*}
where $LSE( \cdot )$ represents a least-square estimation for auto-regressive process introduced in \citep{AutoregressiveImpute2009}.
We also let ${\alpha}_{im, t^*} = 0$, if $E_{\bm{\gamma}}\gamma_{i, t_{near}}<0.5$. Otherwise, we estimate its value in the same way as $\bm{\theta}_{im, t^*}$. 

Then, given the parameter $\bm{\alpha}_{m, t^*}$ and $\bm{\theta}_{m, t^*}$, and the new input point $\bm{x}_{{t}^*}$, the joint distribution of $y_{m}( \bm{x}_{t^*} )$ and observations $\bm{y}_{m}$ can be expressed as:
\begin{align}
\begin{pmatrix} \bm{y}_{m} \\ y_{m}( \bm{x}_{t^*} ) \end{pmatrix} \sim \mathcal{N}
\begin{pmatrix}
	\begin{bmatrix} \bm{\mu} \\ {\mu}_{t^*}
	\end{bmatrix},
	\begin{bmatrix} \bm{\Sigma} & \bm{\Sigma}_{t^*}\\
					\bm{\Sigma}_{t^*}^T & \bm{\Sigma}_{t^*, t^*}
	\end{bmatrix}
\end{pmatrix},
\label{eq:joint distribution}
\end{align}
where ${\mu}_{t^*}=\bm{K}_{(s*)}^T\bm{K}_{(ss)}^{-1}\bm{y}_{(s)}$, $\bm{\Sigma}_{t^*} = \bm{K}_{m*}-\bm{K}_{(sm)}^T\bm{K}_{(ss)}^{-1}\bm{K}_{(s*)}$, and $\bm{\Sigma}_{t^*, t^*} = \bm{K}_{t^*, t^*}-\bm{K}_{(s*)}^T\bm{K}_{(ss)}^{-1}\bm{K}_{(s*)}$.
In the above equations, $\bm{K}_{(s*)}$ is the cross-covariance matrix of the sources and the new input $\bm{x}_{{t}^*}$, $\bm{K}_{(m*)}$ is the covariance of the target observation and the new point, and $\bm{K}_{t^*, t^*}$ is the variance at $\bm{x}_{t^*}$.
Then, the posterior distribution of $y_{m}( \bm{x}_{t^*} )$ can be derived as:
\begin{align}
y_{m}( \bm{x}_{t^*} ) \sim \mathcal{N}  \Big({\mu}_{t^*}+\bm{\Sigma}_{t^*}^T \bm{\Sigma}^{-1} (\bm{y}_m - \bm{\mu}),
\bm{\Sigma}_{t^*, t^*} - \bm{\Sigma}_{t^*}^T \bm{\Sigma}^{-1} \bm{\Sigma}_{t^*} \Big).
\label{eq:posterior distribution}
\end{align}

\section{Numerical study}
\label{sec: numerical study}
In this section, we verify the effectiveness of the proposed non-stationary MGP with the spike-and-slab prior (denoted as DMGP-SS) using synthetic data. In Section \ref{sec: general setting}, we briefly describe the general settings for the numerical study and benchmark methods. In Section \ref{sec: simulation setting}, we introduce the design of simulation cases, where the cross-correlation of the sources and the target are dynamic and sparse. In Section \ref{sec: simulation result}, we demonstrate our model's capability in detecting the underlying correlation pattern as well as improving target prediction performance on the synthetic data.

\subsection{General settings}
\label{sec: general setting}
Similar to the assumption made in the model development section, we generate $m$ sequences consisting of $m-1$ sources and $1$ target, sampled at the same timestamps. The input space is simply time. To investigate the source selection capability of DMGP-SS, we assign only $m_t < m-1$ sources to be correlated with the target at time $t$. Besides, a source will remain either correlated or uncorrelated continuously for a certain period of time.

For comparison, we consider three benchmarks:
\begin{enumerate}
\item
\textbf{GP}. The target is modeled using a single-output GP, with a squared-exponential covariance function.
\item
\textbf{MGP-L1}. It is a state-of-art static method introduced in \citep{RegularizedMGP2022}. MGP-L1 models the target and sources in one MGP model, with the same covariance structure as in \Eqref{eq: covariance matrix}. The scaling parameters $\{\alpha_{im}\}_{i=1}^{m-1}$ are penalized by $L_1$ term to achieve source selection. The regularized log-likelihood of this model is:
\begin{align*}
\log \mathcal{F} = -\frac{1}{2} \bm{y}^T \bm{K}^{-1} \bm{y} - \frac{1}{2} \log |\bm{K}|  - \lambda \sum_{i=1}^{m-1}|\alpha_{im}| - const.
\end{align*}
where $\bm{K}$ is calculated using the static covariance functions in \citep{RegularizedMGP2022} and $\lambda$ is a tunning parameter.  
\item
\textbf{DMGP-GP}. This is a state-of-art non-stationary MGP model, which constructs an LMC model for all outputs and assumes the hyper-parameters follow other GPs \citep{Non-stationaryMGP2021}. More details can be found in the Appendix D. In this model, the covariance for the $m$ outputs is
$${\rm cov}[\bm{y}(x_t),  \bm{y}(x_{t^{\prime}})] = \bm{A}_t \bm{A}_{t'}^T k(x_t, x_{t^{\prime}}) + {\rm diag}\{\sigma_i\},$$
where $\bm{A}_t \bm{A}_{t^{\prime}}^T \in \mathbb{R}^{m \times m}$ is the correlation matrix of $m$ outputs, and ${\rm diag}\{\sigma_i\}$ is the diagonal matrix with  $\{\sigma_i\}_{i=1}^m$. In this study, we focus on the correlation between the sources and target, which corresponds to the last column of $\bm{A}_t \bm{A}_{t}^T$ (except the last element $(\bm{A}_t \bm{A}_{t}^T)_{mm}$), i.e., $(\bm{A}_t \bm{A}_{t}^T)_{0:m-1, m}$.  
\end{enumerate}

All methods are implemented in Python 3.8 and executed on a computer with an Inter(R) Core(TM) i5-7400 CPU with 3.00GHz and 16GB RAM. Both GP and MGP-L1 are implemented using gpflow \citep{GPflow2017}.  DMGP-GP is implemented using TensorFlow and optimized with ADAM \citep{ADAM2014}, with a maximum iteration limit of 500. The EM algorithm for DMGP-SS is also based on Tensorflow, and we use stochastic ADAM with four batches in the M-step. We set the $k_{out}$ and $k_{in}$ in Algorithm \ref{algorithm: summary} to 5 and 400 respectively. 

For MGP-L1, the weight of $L_1$ penalty $\lambda$ is a tuning parameter. For DMGP-GP, we use square exponential functions for $k_{\alpha}$ and $k_{\theta}$. For simplicity, we apply the same tuning parameters for both kernels, i.e., the amplitude ${\alpha}^{\#}$ and length-scale ${\theta}^{\#}$. Those parameters are tuned by cross-validation. In the case of DMGP-SS, the prior sparsity parameter $\eta$ is set to 0.5. We repeat each case 50 times and report the prediction performance through averaging the results. 

\subsection{Simulation cases}
\label{sec: simulation setting}
The main objective of this section is to demonstrate the effectiveness of our method in capturing the non-stationary and sparse cross-correlation between the sources and the target. For simplicity, we hold the other characteristics constant over time, e.g., the smoothness of each output. 
Specifically, we design two simulation cases with different cross-correlation patterns. The first case involves a piece-wise constant cross-correlation, while the second case has a smoothly-changing correlation. In each case, the input data $\{x_t\}_{t=1}^{130}$ are evenly spaced in $[1, 130]$. The observed data are generated from sine functions with measurement noise $\epsilon_t \sim \mathcal{N}(0,0.3^2)$.  

\textbf{Case 1}. 
In this case, we define four kinds of source functions:
\begin{align*}
&Y_{1}(x_t)=3\sin(\pi x_t /20 + e_1) + \epsilon_t,  
\quad Y_{2}(x_t)=2\sin(2 \pi x_t /20 + e_2) \exp[0.5(x_t\%40-1)] + \epsilon_t, \\
&Y_{3}(x_t)=3\sin(4 \pi x_t /20 + e_3) + \epsilon_t, 
\quad Y_{4}(x_t)=2\sin(5 \pi x_t /20 + e_4) + \epsilon_t, 
\end{align*}
where $e_i\sim \mathcal{N}(0,0.2^2)$ is a random phase to make the sampled outputs different from each other, and ``$\%$'' represents the reminder operation. The term $\exp[0.5(x_t\%40-1)]$ is used to deviate the shape of $Y_{2}$ from the standard sine function. In each experiment, we generate $4k$ sources through sampling each kind of function $k$ times, i.e., $m=4k+1$. Specifically, the sources $\{\bm{y}_{i+4j} |  0 \leq j \leq k-1 \}$ are sampled from $Y_i$. 
Then, we define the dynamic target output as: 
$$f_{m}(x_t)=a_{1, t}\sin(\pi x_t/20)+a_{2, t}\sin(2 \pi x_t/20) + a_{3, t}\sin(4\pi x_t/20) + a_{4,t}\sin(5 \pi x_t /20)$$
In this case, we simulate a piece-wise constant cross-correlation by setting: 
\begin{align*}
&a_{1, t} = (2 + 2a_1) I_{t<40}, 
\quad a_{2, t} = (2+ 2a_2) I_{40 \leq t < 80} + (1 + a_2) I_{80 \leq t \leq 130},\\
&a_{3, t} = (1+ a_3) I_{80 \leq t \leq 130}, 
\quad a_{4,t} = 0.
\end{align*}
Therefore, there are three segments, $[0,40)$, $[40, 80)$, and $[80,130]$. Only the $1$st, the $2$nd, and the $2$nd and $3$rd sources are correlated to the target in the three periods, respectively. The other sources remain uncorrelated to the target at all times.

\textbf{Case 2}.
Compared with Case 1, we only change the coefficients $\{a_{i, t}\}_{i=1}^3$ into smoothly-changing ones in this case. Specifically, we let them vary along sine-cosine trajectories, 
\begin{align*}
&a_{1, t} =  [(2+a_1)\cos(\pi t / 120) + 0.5] I_{t<40},\\ 
&a_{2, t} = [(2+a_2)\sin(\pi t / 120 - \pi/6) +0.5] I_{40 \leq t < 130},\\
&a_{3, t} = [(2+a_3) \sin(\pi t / 120 - \pi/2) +0.5] I_{80 \leq t < 130}.
\end{align*}

\begin{figure*}[!t]
\centering
\caption{Dynamic correlation detection results with four sources: (a) Case 1, (b) Case 2. The first column is the true $a_{i,t}$, the second and fourth columns are the estimated ${\bm{\alpha}}_{m}$ for MGP-L1 and DMGP-SS respectively, and the third column shows the estimated $(\bm{A}_t \bm{A}_t^T)_{0:m-1, m}$ for DMGP-GP.}
\subfloat[]{
	\includegraphics[width=6in]{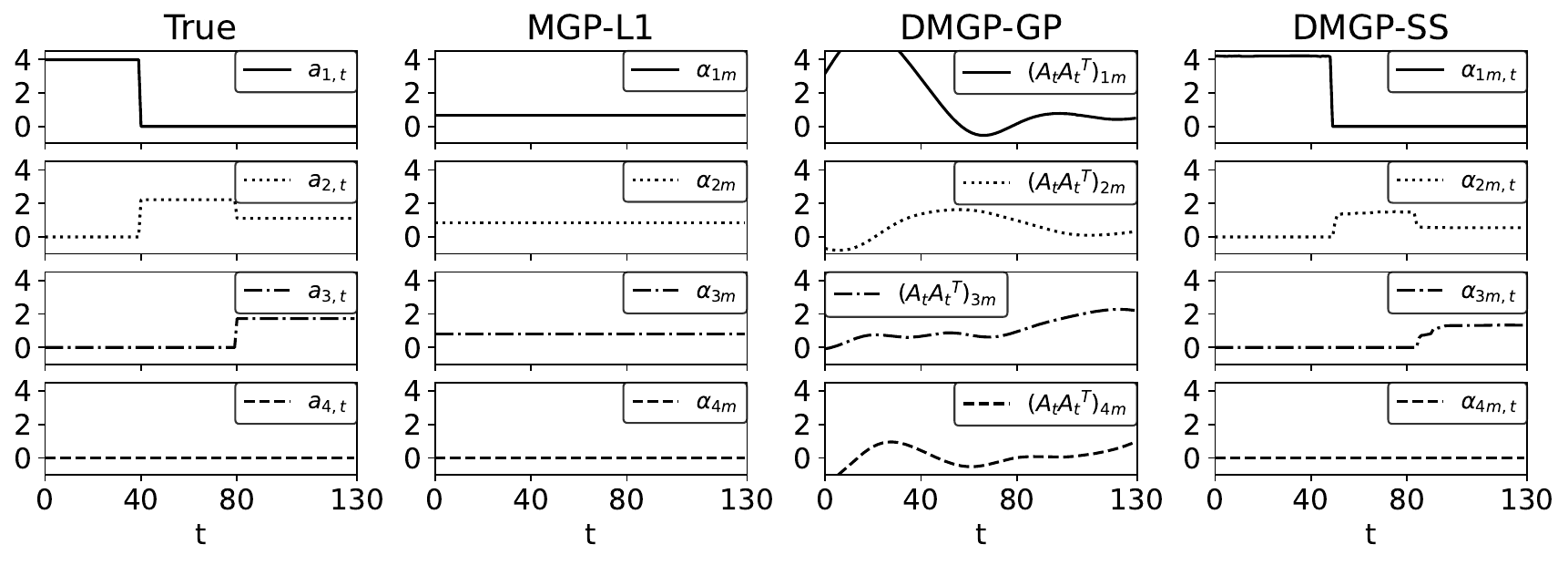}
	\label{fig: case1 scale}
	}\\
\subfloat[]{
	\includegraphics[width=6in]{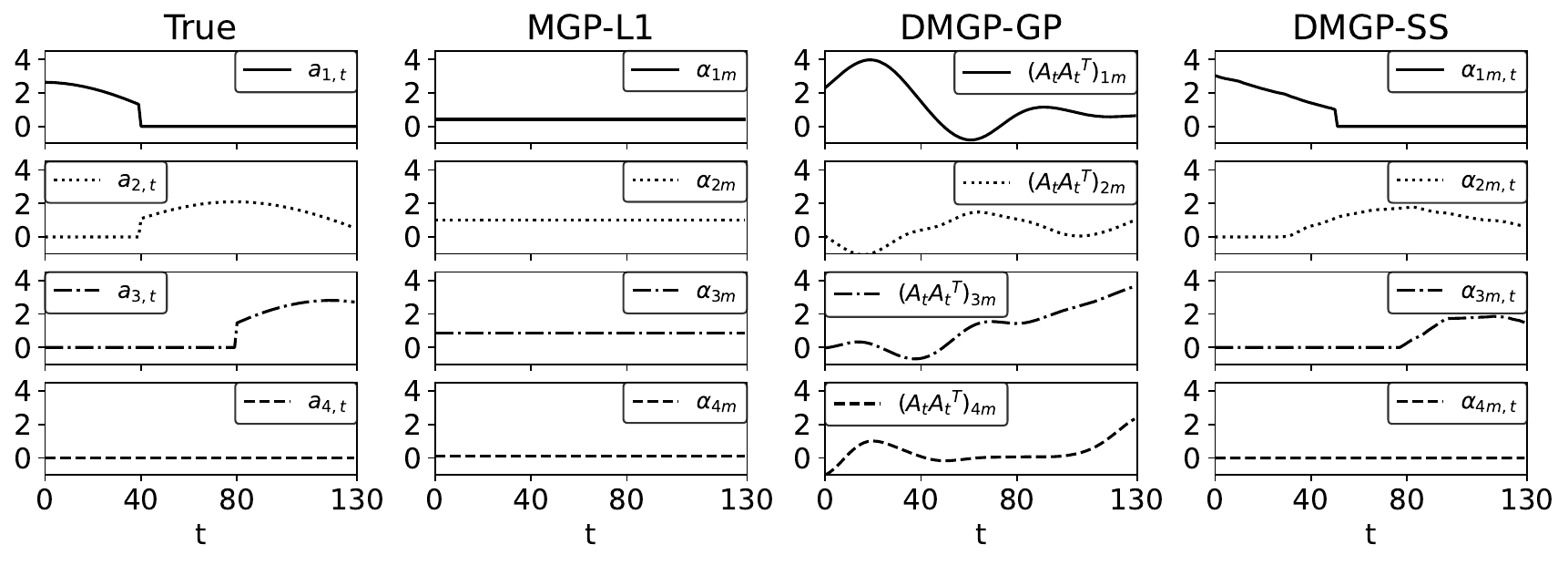}
	\label{fig: case2 scale}
	}
\label{fig: simulation scale}
\end{figure*}

In all cases, we set $k=1, 4$ to generate four and sixteen source outputs for each case. In order to compare the prediction performance of different methods, we randomly remove three short sequences of length $10$ from $[10, 30]$, $[50, 70]$ and $[90, 110]$ respectively. These 30 data points are treated as missing data, while the others are used as training data. 

\subsection{Simulation results}
\label{sec: simulation result}
\begin{figure*}[!t]
\centering
\caption{Visualization of prediction results in Case 1 ($k=1$). The shaded region represents the $99\%$ confidence interval.}
\includegraphics[width=6in]{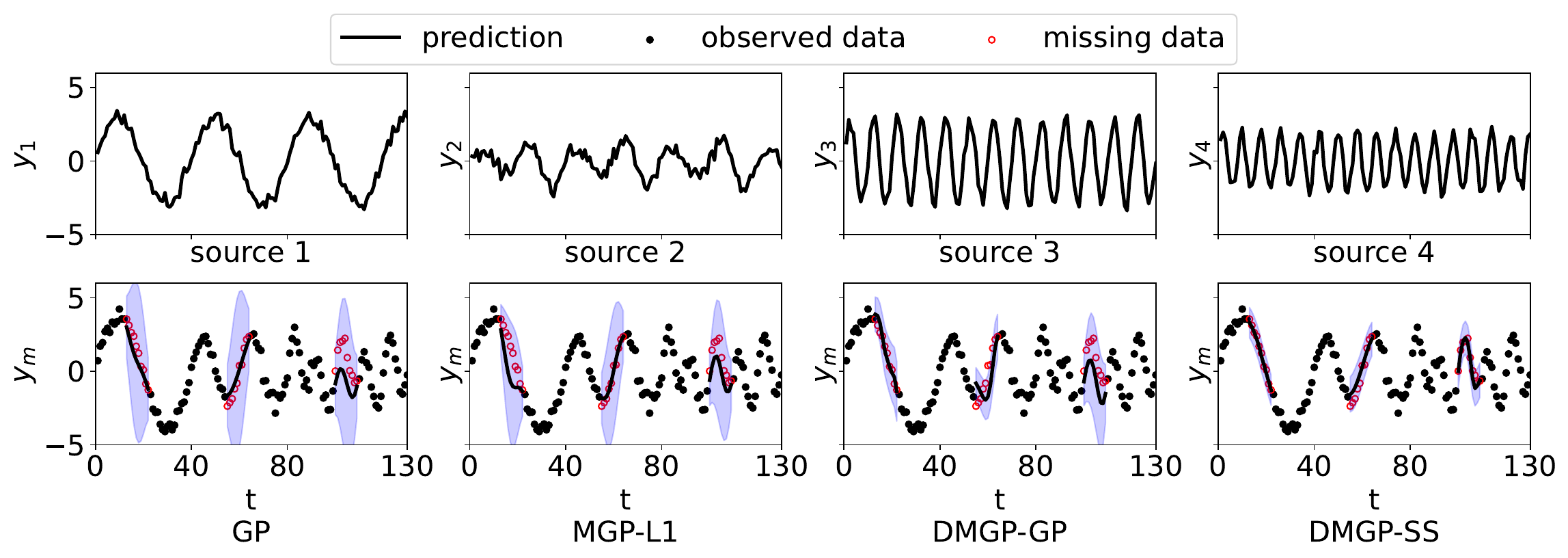}
\label{fig: case1 prediction}
\end{figure*}

To begin with, we demonstrate that the proposed DMGP-SS is capable of capturing the dynamic and sparse correlation between the sources and the target. 
\Figref{fig: simulation scale} illustrates the estimated $\{{\alpha}_{im}\}_{i=1}^4$ for MGP-L1, the $\{{\alpha}_{im, t}\}_{i=1}^4$ for DMGP-SS, and the estimated $(\bm{A}_t \bm{A}_t^T)_{1:4, m}$ for DMGP-GP in Case 1 and 2 with four sources. And \Figref{fig: case1 prediction} visualizes the sources and target prediction results in Case 1 with four sources. DMGP-SS is implemented with the hard and soft slab priors for Case 1 and 2 respectively.  Note that the value of $a_{i,t}$ and $\alpha_{im, t}$ are not identical, since $a_{i,t}$ is a linear combination weight rather than the correlation parameter in MGP.  

Overall, DMGP-SS successfully recovers the true dynamic structural sparsity of correlation shown in the first column of \Figref{fig: simulation scale}. Firstly, DMGP-SS tracks closely the periods of correlation between each source and the target, achieving a dynamic selection of sources. Since the target's characteristics do not change abruptly (as shown in \Figref{fig: case1 prediction}), it is reasonable that the estimated correlation change points are about ten time-steps before or after the designed change times. Second, the hard slab prior encourages nearly piece-wise correlation, while the soft slab prior allows smoothly changing correlation. \textcolor{black}{Due to the appropriate selection of sources at different times in \Figref{fig: case1 prediction}, the proposed DMGP-SS achieves precise prediction with the lowest prediction uncertainty. This highly improves the confidence of decision making when using the recovered series. The difference on confidence interval of three MGP models is due to that the uncorrelated sources 'poison' the correlation structure and decrease the influence of the truly-correlated sources, resulting in a lower value of variance reduction term $\bm{\Sigma}_{t,*}^T \bm{\Sigma}^{-1} \bm{\Sigma}_{t,*}$ in posterior prediction \Eqref{eq:posterior distribution}.}

In contrast, another non-stationary method DMGP-GP fails to estimate a sparse structure in the source-target correlation since the GP prior on parameters lacks the shrinkage effect. The proposed DMGP-SS addresses this problem through combining the smooth slab prior and the shrinking spike prior.
Regarding MGP-L1, although it can estimate a sparse structure of $\bm{\alpha}_{m}$, the estimated values of non-zero parameters are constant over time and cannot reflect the change of correlation. As a result, it even performs worse than GP in recovering the target output in  \Figref{fig: case1 prediction}.

\begin{figure}[!t]
\centering
\caption{The estimated $E_{\bm{\gamma}}{\bm{\gamma}}_{1:4}$ for DMGP-SS in Case 1.}
\includegraphics[width=6.0in]{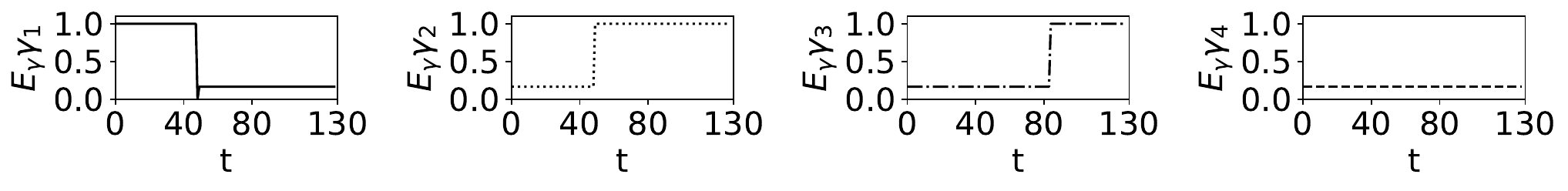}
\label{fig: simulation gamma}
\end{figure}

Another advantage of DMGP-SS is the adaptive adjustment of the spike-and-slab combination weight, $E_{\bm{\gamma}}{\gamma}_{i, t}$. In our settings, $\nu_0^{-1}$ in the spike prior is much larger than $\nu_1^{-1}$ in the hard slab prior (or $\sqrt{\nu_1^{-1}}$ in the soft slab prior), to put more penalty on the correlation sparsity. For example, we set $\nu_0 = 0.02$ and $\nu_1 = 0.1$ in Case 1. However, this sparse penalty does not cause significant bias on non-zero $\alpha_{im, t}$ because of the automatically adjusted $E_{\bm{\gamma}}{\gamma}_{i, t}$ in the EM algorithm. Specifically, we initialize it with $0.99$ to barely shrink parameters at the beginning. Then $E_{\bm{\gamma}}{\gamma}_{i, t}$ is updated in the E-step of the EM algorithm. \Figref{fig: simulation gamma} shows its estimated value after five iterations. For the correlated sources (e.g., the first source during $t\in[0,50]$), their corresponding $E_{\bm{\gamma}}{\gamma}_{i, t}$ is very close to $1$, so they bear negligible shrinkage effect from the spike prior. In contrast, for the uncorrelated sources (e.g., the first source after $t = 50$), $E_{\bm{\gamma}}{\gamma}_{i, t}$ is approximately $0.2$, implying a substantial shrinkage effect. The value $0.2$ can be derived based on \Eqref{eq: gamma update}. For consecutive zero elements, $p_{slab}=\eta (2\nu_1)^{-1}$, and $p_{spike}=(1-\eta) (2\nu_0)^{-1}$, resulting in $E_{\bm{\gamma}}{\gamma}_{i, t} \approx \nu_1^{-1}/\nu_0^{-1}$.

\begin{table}[!t] 
\renewcommand{\arraystretch}{1.0}
\setlength\tabcolsep{3pt}
\setlength\abovecaptionskip{0cm}  
\caption{Prediction error for Case 1 and 2. Values in the brackets are standard deviations.}
\label{tab: simulation MAE}
\centering
\begin{tabular}{c | c c c c | c c c c}
\hline
\multirow{2}{*} {outputs}  & \multicolumn{4}{c|}{Case 1} & \multicolumn{4}{c}{Case 2} \\
\cline{2-9}
& GP  & MGP-L1 & DMGP-GP & DMGP-SS  & GP  & MGP-L1 & DMGP-GP & DMGP-SS   \\
\hline
  \multirow{2}{*}{5} & 0.99 & 0.89  & 0.76 & 0.56  & 1.12 & 1.04  & 0.85 & 0.66   \\
&(0.26)& (0.21)  & (0.23)& (0.13) &(0.27)& (0.24)  & (0.26)& (0.17)   \\
 \multirow{2}{*}{17} & 0.95 & 0.91  & 0.75 & 0.50 & 1.15 & 1.02  & 0.78 & 0.64  \\
&(0.25)& (0.21)  & (0.19)& (0.15) &(0.26)& (0.21)  & (0.17)& (0.16)  \\
\hline
\end{tabular}
\end{table}

\Tabref{tab: simulation MAE} summarizes the results of 40 repetitions for each case.  
DMGP-SS outperforms all the other methods in both cases. And the increasing of source number does not affect the prediction accuracy,   demonstrating its remarkable robustness in dealing with moderate data. 
MGP-L1 exhibits slightly higher prediction accuracy than GP. Because the static covariance structure limits its ability to transfer accurate information at all times. Under some circumstances, this limitation will cause a negative transfer effect on the learning of target (for example, the result shown in \Figref{fig: case1 prediction}).
DMGP-GP has a better performance on target prediction than GP and MGP-L1, due to the ability to model dynamic correlation. However, it does not achieve the same prediction accuracy as DMGP-SS in these cases. On the one hand, it lacks the ability to exclude the impact of uncorrelated sources. On the other hand, DMGP-GP is an extension of LMC and uses the same function $k(x_t, x_{t^{\prime}})$ to model the auto-covariance of every output. This feature makes it unsuitable for our cases where the source covariance functions have four kinds of length-scales. Nevertheless, the proposed DMGP-SS models each source with separate kernels and latent functions, highly increasing its flexibility.

{\color{black} \Tabref{tab: computational time} lists the computational time of the four methods in Case 1. Between the two non-stationary methods, DMGP-SS requires much less computation time than DMGP-GP. This exactly verifies the analysis in \ref{subsec: computational challenge} that our model can save a large mount of time than the traditional non-stationary MGP method, due to a block-sparse covariance matrix. In fact, our model is scalable for larger size of data, which is described in Appendix E.
\begin{table}[H]
    \centering
    \caption{Computational time for the different methods in Case 1.}
    \begin{tabular}{cc|cccc}
    \hline
         outputs & n & GP & MGP-L1 & DMGP-GP & DMGP-SS\\
         \hline
         5 & 130 & 0.13 & 8.6 & 130 & 73 \\
         17 & 130 & 0.10 & 26 & 2100 & 300 \\
    \hline
    \end{tabular}
    \label{tab: computational time}
\end{table}}

\section{Case study}
\label{sec: case study}

In this section, we apply DMGP-SS to two cases: human movement signal modeling and control policy optimization. In the first case, these signals are recorded by sensors attached to different joints, such as hands and feet. As the movement of joints are dynamically correlated across different gestures \citep{DynamicSubspace2022}, it is reasonable to utilize a non-stationary method to capture the varying correlation and leverage information among the signals of joints. Regarding the control policy iteration, we study on a classical reinforcement learning problem, mountain-car. We evaluate the performance of DMGP-SS on leveraging knowledge between difference systems when the environment is non-stationary. 

\subsection{Movement Signal Modeling}
\subsubsection{Data Description}
We use the MSRC-12 gesture dataset \citep{GestureData2012} consisting of sequences of human skeletal body movement. Twenty sensors are distributed across the body, and each sensor can record three-dimensional coordinates of joint positions. The motion signal has a sampling frequency of 30 Hz and an accuracy of $\pm 2cm$. The dataset comprises 12 different gestures performed by 30 participants. The experiment involves each participant starting from a standing position, performing one gesture, and returning to the standing state. This process repeats several times continuously within each sample. 

\begin{figure*}[!t]
\centering
\caption{(a). The snapshots of ``Stand'', ``Google'', ``Shoot'' and ``Throw''. The twenty joints are represented by circles. We mark out the four selected joints for our study, where ``L'' and ``R'' represent ``left-side'' and ``right-side'' respectively, and ``E'' and ``W'' represent the elbow and wrist joints respectively. (b). The twelve movement signals of the selected joints in the three gestures' data, where the red signal is takes as the target signal and the others are the source signals. In the label of vertical axis, ``x'', ``y'' and ``z'' represent different coordinates. Each signal has a vertical range of $[-2,2]$, with the horizontal ticks representing time indexes.}
\subfloat[]{
	\includegraphics[width=1.5in]{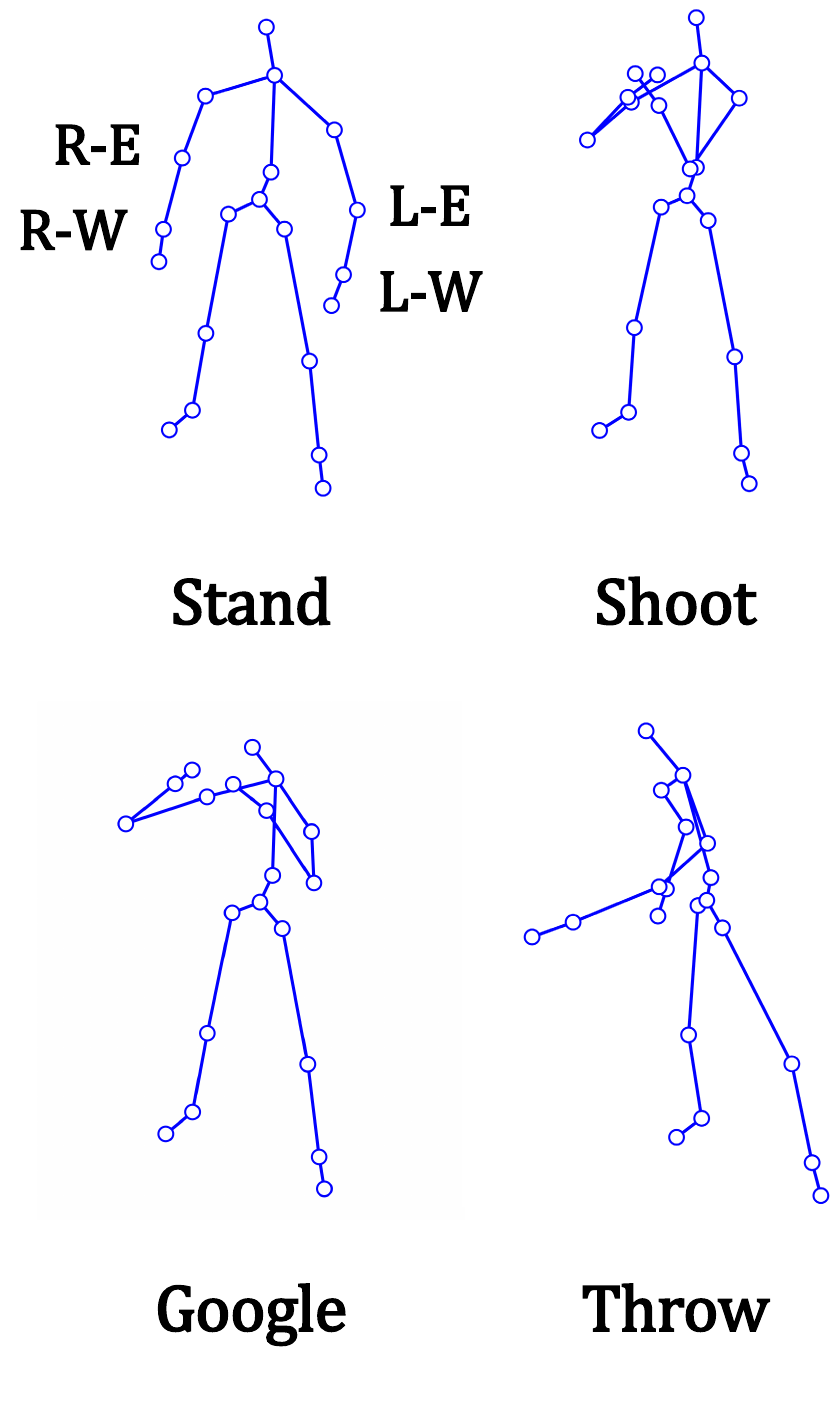}
	\label{fig: gesture skl}
	}
\subfloat[]{
	\includegraphics[width=3.5in]{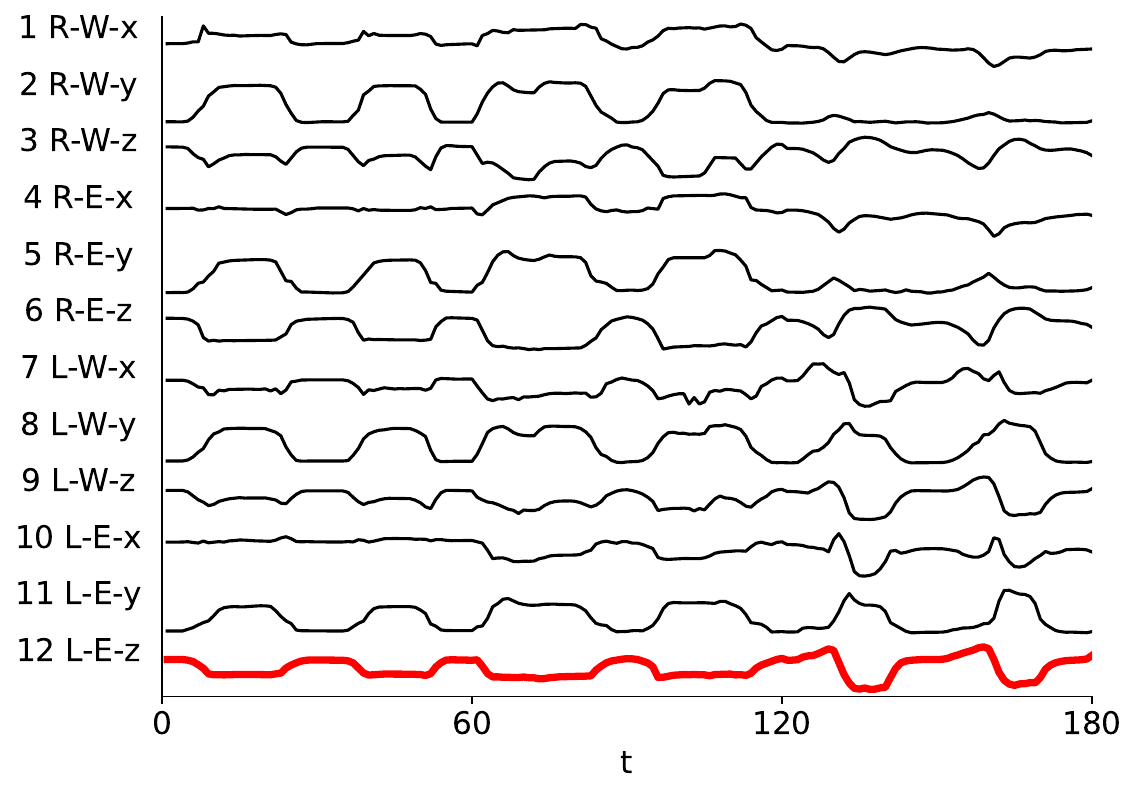}
	\label{fig: real data}
	}
\label{fig: simulation scale}
\end{figure*}

To demonstrate the effectiveness of DMGP-SS, we connect the instances of three gestures (``Goggles'', ``Shoot'', and ``Throw'') performed by the same individual.  \Figref{fig: gesture skl} shows the snapshots of the standing position and the selected gestures.
In the first two gestures, the participant stretches both arms in front of him to perform searching or shooting motions. In the third gesture, the participant only uses his left arm to make an overarm throwing movement. In these gestures, the main acting joints are hands, wrists, and elbows, where the trajectories of hands and wrists are almost identical. Therefore, we select the movement signals of four joints (left wrist, left elbow, right wrist and right elbow) as twelve outputs. We choose the $z$-coordinate movement of the left elbow as a target output, while using the remaining eleven movement signals as sources.

We select two 120-frame-long instances for each gesture and down-sample each instance to 30 frames. Therefore, there are 180 points for each output. To eliminate the difference in initial coordinate values, we reset the initial three-dimensional coordinate value to $(0, 0, 2)$ across different recordings. Additionally, we re-scale all outputs to $[-2, 2]$.  \Figref{fig: real data} displays the 12 outputs and the change of joints' positions.

\subsubsection{Results}
Intuitively, the cross-correlation between the source and target signals should remain constant for a single gesture, so a hard slab prior is used in DMGP-SS. All other settings for this case are identical to those used in the simulation studies. We still simulate consecutive data missing for the target. From the 60-points-long time-series of each gesture, we randomly remove a sequence of length ten as missing data.

\begin{figure*}[!t]
\centering
\caption{(a). Correlation estimation results for the data of three gestures. The first and third row are the estimated $\bm{\alpha}_{m}$ for MGP-L1 and DMGP-SS respectively, and the second row shows the estimated $\{\bm{A}_t \bm{A}_t^T\}_{1:11, 12}$ for DMGP-SS. (b). Recovery results of four methods for the movement signal of right wrist.  The shaded region represents the $99\%$ confidence interval.}
\subfloat[]{
	\includegraphics[width=3in]{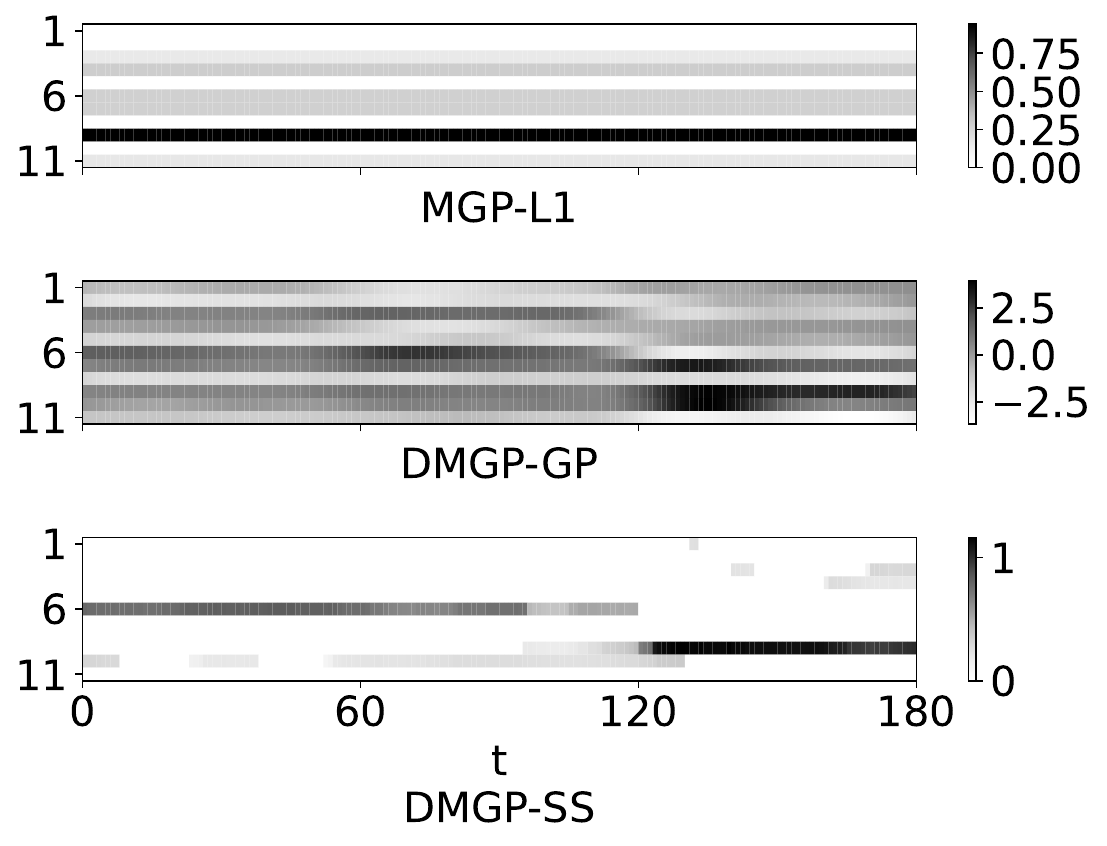}
	\label{fig: real correlation}
	}
\subfloat[]{
	\includegraphics[width=3in]{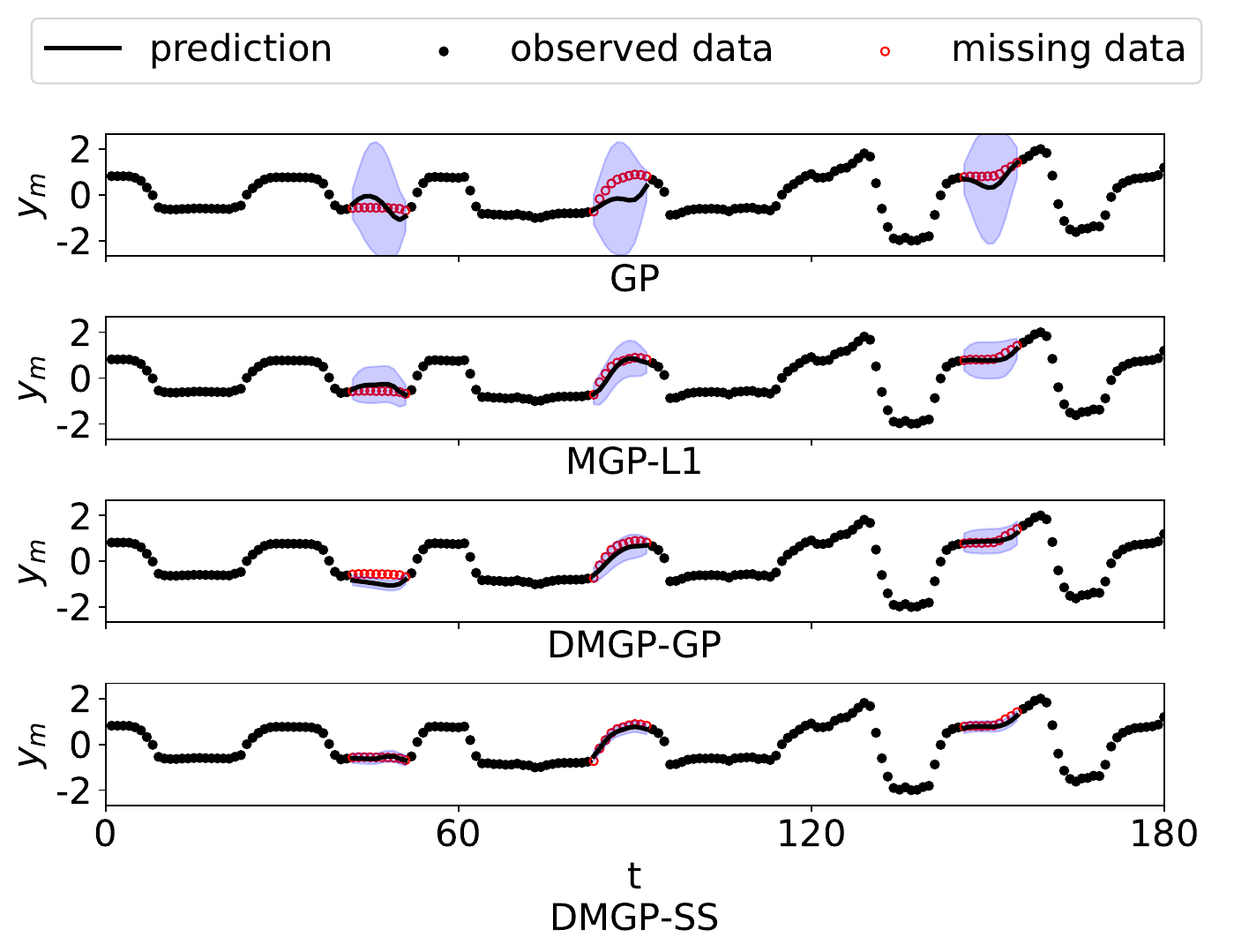}
	\label{fig: real prediction}
	}
\label{fig: simulation scale}
\end{figure*}

First of all, \Figref{fig: real correlation} shows the estimated correlation between the sources and the target. MGP-L1 selects six sources (the third 'R-W-z', the forth 'R-E-x', the sixth 'R-E-z', the seventh 'L-W-x', the ninth 'L-W-z', and the eleventh 'L-E-y')  as the correlated signals for the whole time period, in which the ninth source has the strongest correlation. DMGP-SS selects the sixth source (`R-E-z') and the ninth source (`L-W-z') as correlated sources when $0 \leq t \leq 120$ and $120 \leq t \leq 180$, respectively. DMGP-GP does not provide a sparse estimation of cross-correlation. Among them, the proposed DMGP-SS accurately captures the underlying physical movements of each gesture. In the ``Google'' and ``Shoot'' gestures, both elbows have almost the same trajectory. In the ``throw'' gesture, the left elbow's movement is highly correlated with that of the left wrist. These findings align well with the signals shown in \Figref{fig: real data}. On the contrary, limited by the static correlation structure, MGP-L1 can only select some sources and force them to be correlated with the target all the time, but such signals do not exist in the dynamic environment. For DMGP-GP, the results cannot provide us an intuitive understanding on the joints' relationship.

%

\Figref{fig: real prediction} displays the recovered target signal in one experiment. Notably, DMGP-SS accurately recovers the missing data with high precision. Besides, it has the minimal uncertainty because it can selects the most correlated sources at each time and such a high correlation improves the confidence on prediction results.  Conversely, the predictions of MGP-L1 and DMGP-GP display an obvious bias in the first gap and higher uncertainty for all gaps. \textcolor{black}{Besides, similar to the results of numerical studies, the proposed DMGP-SS also gives predictions with the lowest uncertainty, which significantly improves the confidence of decision making.}

{\color{black} We further repeat the experiments 36 times. \autoref{tab: real MAE} compares the prediction accuracy of four methods in terms of both the MAE and the continuous-ranked-probability-score (CRPS). The CRPS measure is a widely used metric to evaluate the probabilistic forecast for a real-valued outcome \citep{CRPS2000}: 
$$CRPS = {n_{test}^{-1}} \sum_{i=1}^{n_{test}} \int \left[\Phi \left( \hat{y}_i \right) - 1_{\hat{y}_i \geq y_i)} \right]^2 d\hat{y}_i $$
where $\hat{y}_i$ is the predicted output and $\Phi$ is the cumulative density function (CDF) of the predicted distribution. A low CRPS value suggests that the predicted posterior are concentrated around the true value, indicating a high probabilistic forecast performance.} As expected, DMGP-SS outperforms the other methods due to its ability to capture the dynamic and sparse correlation accurately. MGP-L1 performs better than GP, benefiting from the information borrowed from the other sources. However, it cannot model a dynamic correlation, resulting in lower prediction accuracy than DMGP-SS. Regarding DMGP-GP, although it captures the change of correlations between the target and some sources, its prediction accuracy is even lower than GP. This result may be attributed to that non-sparse correlations lead to potential negative transfer effects. Besides, since the sources' smoothness is heterogeneous, it is inappropriate to use the same auto-covariance function to model all outputs. 

\begin{table}[!t] 
\renewcommand{\arraystretch}{1.0}
\setlength\tabcolsep{5pt}
\setlength\abovecaptionskip{0cm}  
\caption{Prediction error for real case. The values in the bracket are standard deviations.}
\label{tab: real MAE}
\centering
\begin{tabular}{c| cccc}
\hline
 & \makecell*[c]{GP} &  \makecell*[c]{MGP-L1 }  & \makecell*[c]{DMGP-GP} &  \makecell*[c]{DMGP-SS} \\
\hline
MAE-mean & 0.43 & 0.22 & 0.50 & 0.18   \\
MAE-std. & (0.12) & (0.10) & (0.25) & (0.08) \\
CRPS-mean & 0.30 & 0.16 & 0.41 & 0.15 \\
\hline
\end{tabular}
\end{table}

\subsection{Control Policy Optimization}
\color{black}
In reinforcement learning problems, there are five important elements for the agent: state $s$, action $a$, reward $r$, value $V$ and policy $W$. Starting from one state $s$, the agent takes an action $a$, transitions into a new state $u$ and gets an immediate reward $r(u)$ from the environment. In general, a reinforcement learning framework includes two primary procedures: 
\begin{enumerate}
    \item Model estimation: estimating the state transition function $U(s, a)$ based on the observed transition samples $[(s, a), u]$.
    \item Policy iteration: iteratively estimating the state value $V(s)$ (the long-term expected reward) for each state and improving the control policy $W(s)$.
\end{enumerate}
More details could be found in the works on GP-based reinforcement learning \citep{GPRL, LMCRL}. 

\begin{figure}[!t]
\centering
\caption{The illustration of mountain-car problem.}
\includegraphics[width=3.0in]{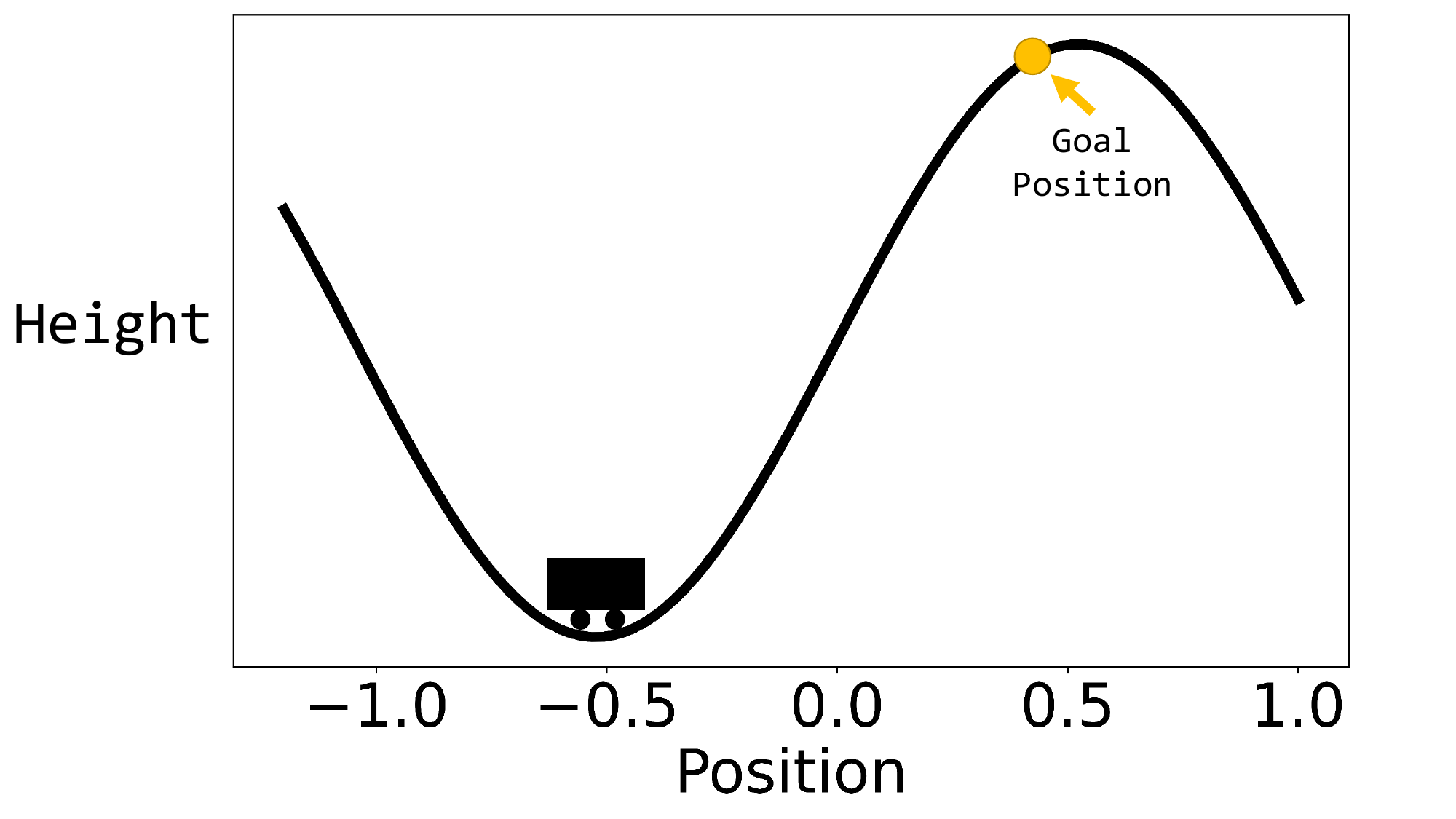}
\label{fig: mountain car}
\end{figure}

In this study, we employ the well-known reinforcement learning problem, mountain-car, to demonstrate the application of our model in decision making. \Figref{fig: mountain car} illustrates such a problem. A car begins in a valley and aims to reach a goal position of the right-most hill. Due to the steep slope of the hill, the car cannot directly accelerate to the goal position. Instead, it needs to drive up the opposite side, turn back, and accelerate to reach the goal position. In system, the agent state is described by the position $s^{\text{pos}}\in[-1.2, 1.0]$ and the velocity $s^{\text{vel}}\in[-0.07, 0.07]$ of the car, i.e., $\bm{s}=(s^{\text{pos}}, s^{\text{vel}})$. The agent action is a horizontal force $a \in[-1, 1]$. The car starts at the state $\bm{s}_{\text{init}}=(s^{\text{vel}}_{\text{init}}, s^{\text{vel}}_{\text{init}})$ with $s^{\text{pos}}_{\text{init}} \in [-0.6, -0.5]$ and $s^{\text{vel}}_{\text{init}}=0$, aiming to reach the goal state $\bm{s}_{\text{goal}}=(0.45, 0)$. The reward function is the probability density function of $N(\bm{s}_{\text{goal}}, \text{diag}\{0.05^2, 0.0035^2\})$. The dynamic equation of this system is approximated by:
\begin{align}
    & s^{\text{vel}}_t = s^{\text{vel}}_{t-1} + P \cdot a_{t-1} - G \cdot \cos(3 \cdot s^{\text{pos}}_{t-1}) \notag\\
    & s^{\text{pos}}_t = s^{\text{pos}}_{t-1} + s^{\text{vel}}_{t} 
    \label{eq: true transition}
\end{align}
where $P$ is the horizon power unit and $G$ is the vertical force unit.

We consider the control problem involving a car in a non-stationary environment with limited transition samples. During $t\in[0,20)$, the target car runs in an environment with $(P,G)=(0.009, 0.0015)$ and we get 20 random samples. However, at $t=20$, an unknown change occurs, altering  the environment factors to $(P, G)=(0.0011, 0.0026)$. We sampled another 20 random samples during $t\in [20,40)$. After $t=40$, we start to control the car to reach the goal position and the environment does not change any more. Given that those samples are too limited to build an accurate a transition model, we transfer information from two historical source datasets, each consisting of $200$ samples from stationary environments with $(P,G)=(0.01, 0.0015)$ and $(P, G)=(0.001, 0.0025)$, respectively. We can see that the target environment is close to the first source environment before $t=20$, and is close to the second one after that.


Algorithm \ref{algorithm: offline mountain car} summarizes the workflow of reinforcement learning, where we employ the DMGP-SS as a transition model. For simplicity, we model the source transition data with stationary GPs in DMGP-SS. The sources and the target are expressed as:
\begin{align}
u_{i} (\bm{s}, a) & =\alpha_{ii} g_{ii}(\bm{s}, a)\ast z_{i}(\bm{s}, a) + \epsilon_i , i \in \mathcal{I}^S, \notag\\
u_{m} (\bm{s}, a) & = \sum_{j \in \mathcal{I}}\alpha_{jm, t}g_{jm, t}(\bm{s}, a) \ast z_{j}(\bm{s}, a) + \epsilon_m.  
\label{eq: dynamic MGP for RL}
\end{align}
Here, $u$ is the position or velocity state to where the agent transitioned from a state $\bm{s}$ after taking an action $a$. After fitting this model both the target and source samples, we train a GP for the value model $V(\bm{s})$ with even-spaced support points ${S}_{\text{supp}}$ and rewards $r({S}_{\text{supp}})$. Based on the transition model $U(\bm{s}, \bm{a})$, we iteratively improve the policy $W(\bm{s})$ and value model $V(\bm{s})$ until the prediction of $V(\bm{s})$ converges. Details on the policy improvement procedure can be found in \citep{GPRL, LMCRL}. While we focus on the offline setting in this case, it is worth noting that this framework can be readily extended to accommodate an online setting, wherein the training sample consists of the visited states and the transition models are updated every few steps.
\begin{algorithm}
\caption{Control Policy Optimization for Mountain Car Case}
\label{algorithm: offline mountain car}
\begin{algorithmic}[1]
\Require Source transition samples $\{ (\bm{s}_{i,n}, a_{i,n}), \bm{u}_{i, n}\}_{i,n=1}^{2,200}$, target transition samples $\{(\bm{s}_{3,t}, a_{3,t}), \bm{u}_{3,t}\}_{t=1}^{40}$, reward function $\{r(\bm{s})\}$, 256 support points ${S}_{\text{supp}}$.  
\State Initialize: policy $W(\bm{s}) \leftarrow$ random policy, value ${V}_{\text{supp}} = r({S}_{\text{supp}})$. 
\State Fit two DMGP-SS models for the state transition using both the source and the target transition samples, one for the position state and the other for the velocity state.
\State Fit a GP for the state value model $V(\bm{s})$ using $\{{S}_{\text{supp}}, {V}_{\text{supp}}\}$.
\State Improve policy $W(\bm{s})$ and value model $V(\bm{s})$ iteratively based on the state-transition model until $V({S}_{\text{supp}})$ converges.
\State Execute the optimized policy starting at the state $\bm{s}_{\text{init}}$:
\end{algorithmic}    
\end{algorithm}

\begin{figure*}[!t]
\centering
\caption{The control path of the three methods: (a) RL-GP (b) RL-MGP, and (c) RL-DMGP-SS. The blue points mark the positions every 50 steps.}
\subfloat[RL-GP]{
	\includegraphics[width=2in]{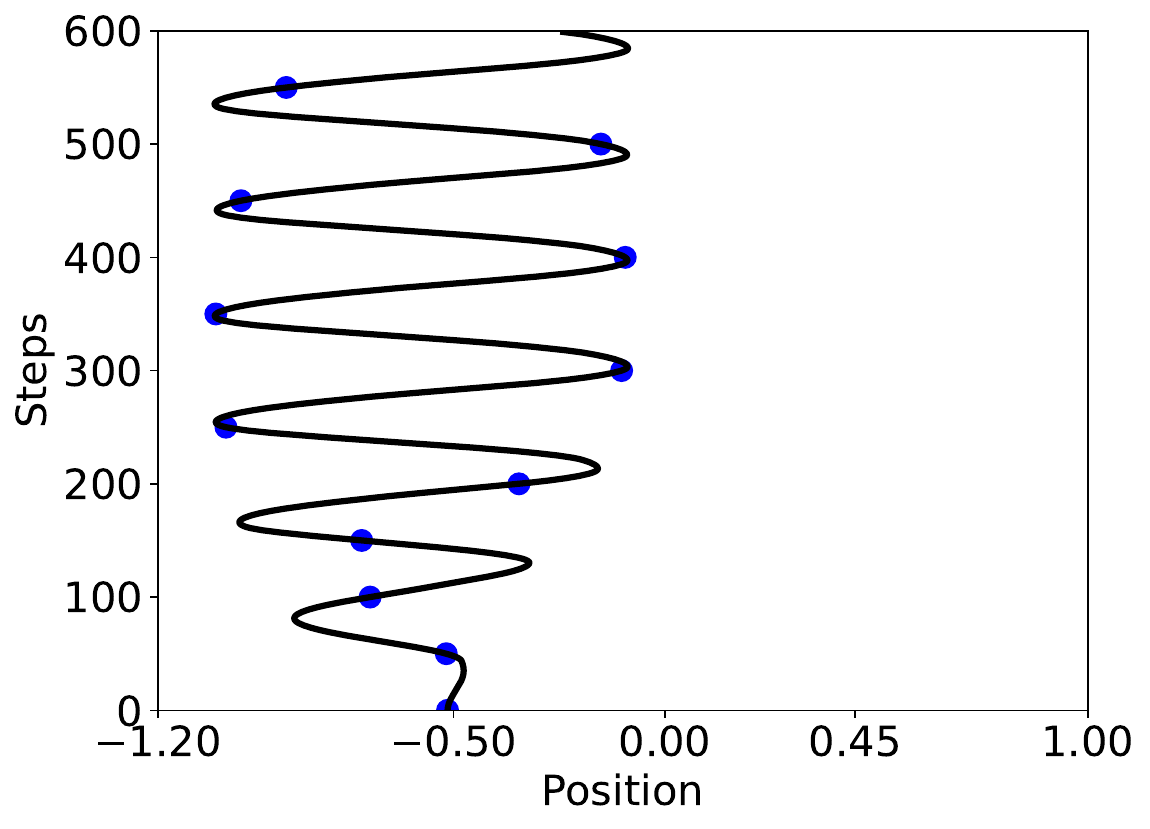}
	\label{fig: RL-GP path}
	}
\subfloat[RL-MGP]{
	\includegraphics[width=2in]{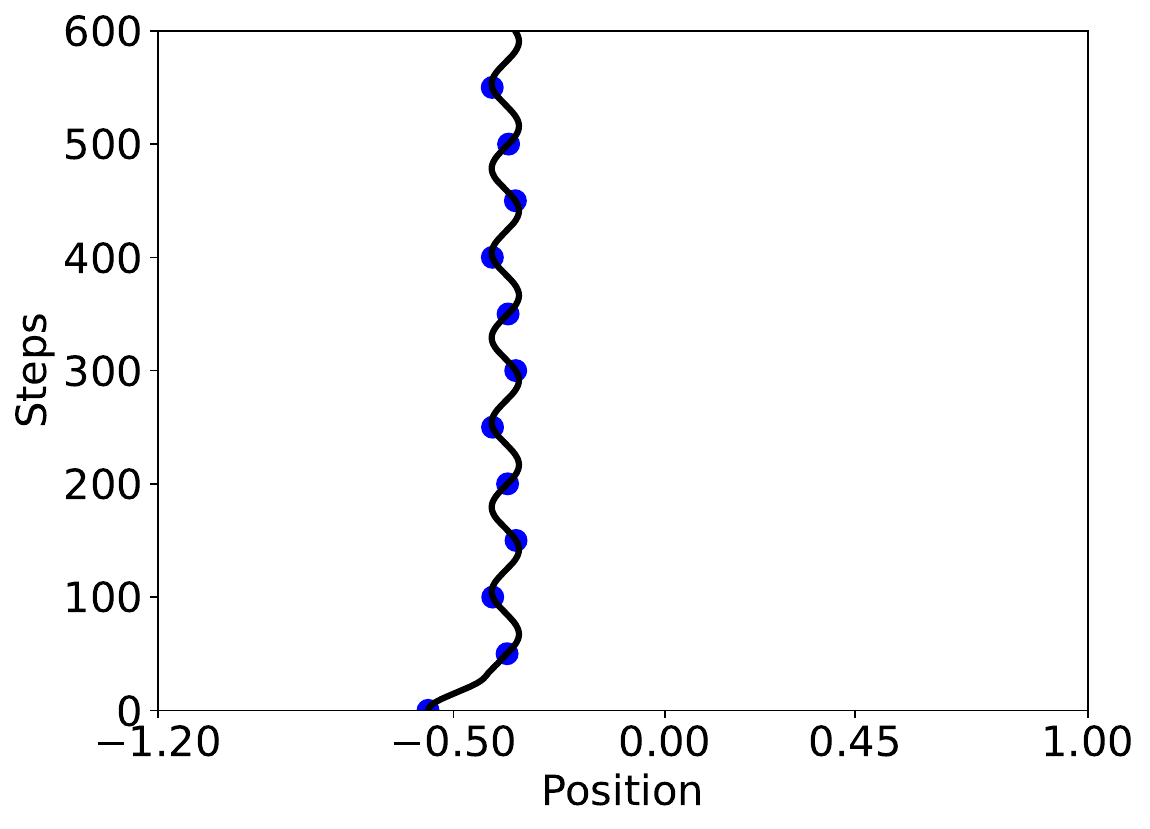}
	\label{fig: RL-MGP prediction}
	}
 \subfloat[RL-DMGP-SS]{
	\includegraphics[width=2in]{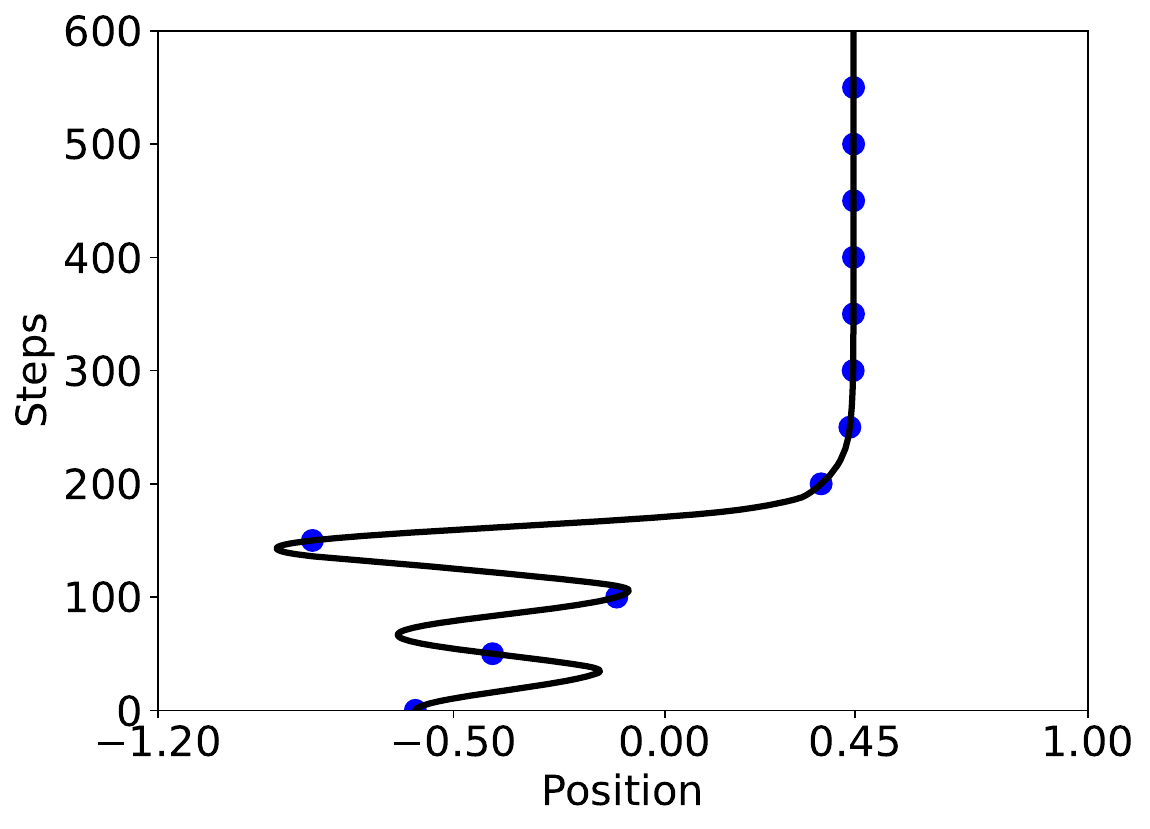}
	\label{fig: RL-DMGP-SS prediction}
	}
\label{fig: rl case path}
\end{figure*}

We choose two reinforcement learning benchmark methods \citep{GPRL, LMCRL} with the stationary GP and MGP as the state transition model, respectively. The maximum execution steps are $600$ for each method. 
In \Tabref{tab: RL MAE and Distance}, we report the the mean of absolute distances to the goal state for three methods. The DMGP-SS-based control policy has the shortest average distance to the goal state in 600 steps. Specifically, \Figref{fig: rl case path} compares the position of the car controlled by the three policies. With DMGP-SS as the transition model, the car reaches the goal position and stays there after about 250 time steps. However, the other methods cannot find a good policy to reach the goal position within 600 moves, since their stationary transition models cannot account for the environment change and make the policy iteration hard to converge.

\begin{table}[!t] 
\renewcommand{\arraystretch}{1.0}
\setlength\tabcolsep{5pt}
\setlength\abovecaptionskip{0cm}  
\caption{Predictive accuracy on state transition and the mean of absolute distances to the goal state in the mountain-car case.}
\label{tab: RL MAE and Distance}
\centering
\begin{tabular}{c| ccc}
\hline
Metric & \makecell*[c]{RL-GP} &  \makecell*[c]{RL-MGP}  &  \makecell*[c]{RL-DMGP-SS} \\
\hline
\makecell*[c]{The mean of absolute distances to \\ the goal position} & 0.98 & 0.83 & 0.27 \\
\makecell*[c]{Predictive MAE on velocity  \\ transition ($10^{-3}$)} & 3.8 & 5.8 & 0.51   \\
\makecell*[c]{Predictive MAE on position \\ transition ($10^{-3}$)} & 3.6 & 4.9 & 8.2 \\
\hline
\end{tabular}
\end{table}

\begin{figure}[!t]
\centering
\caption{The estimated correlation parameter $\bm{\alpha}_m$ from DMGP-SS on velocity transition.}
\includegraphics[width=3.0in]{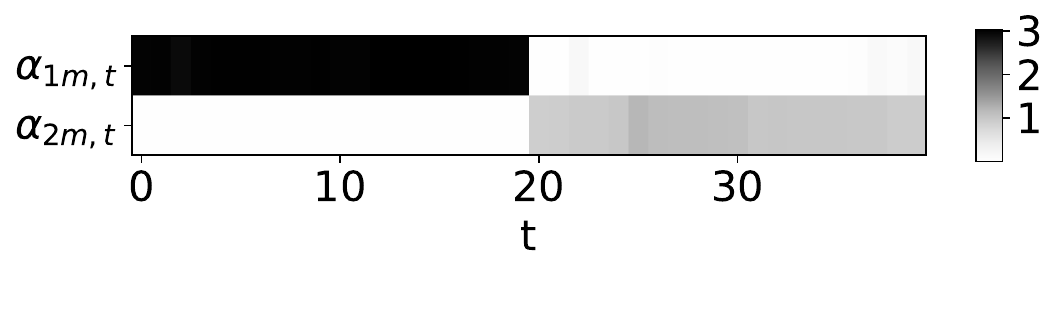}
\label{fig: rl correlation}
\end{figure}

In \Tabref{tab: RL MAE and Distance}, we further compare the predictive MAE on state transition for three methods. The proposed method has a significantly lower prediction error on velocity state transition than the the other methods, since it can capture the change of environment and transfer information from the related sources. \Figref{fig: rl correlation} illustrates the estimated $\bm{\alpha}_m$ from DMGP-SS trained on the velocity transition data. We can find that the proposed method successfully finds that the correlation between the sources and the target changes at $t=20$. Therefore, during the policy improvement stage, it can leverage information from the similar source (the second one) and avoid the negative transfer from the uncorrelated source (the first one). Regarding the prediction error on position transition, although RL-DMGP-SS has a higher MAE than the other benchmarks, the difference is minor considering the position range $[-1.2, 0.6]$. Therefore, our method can provide the best control policy.

\color{black}

\section{Conclusion}
\label{sec: conclusion}
This paper proposes a flexible non-stationary multi-output Gaussian process for modeling multivariate data in transfer learning. The novelty of our approach lies in its ability to capture dynamic and sparse cross-correlations between sources and targets. We achieve this by allowing correlation parameters to follow a spike-and-slab prior, where the slab prior ensures correlation variation over time, and the spike prior encourages parameters to shrink to zero, eliminating negative transfer effects from uncorrelated sources. The ratio of these two priors is automatically adjusted in the proposed EM algorithm, preventing the shrinkage effect on non-zero correlation parameters. Through the experiments on simulation and human gesture dataset, we demonstrate that our method is well-suited for both capturing non-stationary characteristics and mitigating negative transfer. 

\textcolor{black}{The proposed data-driven method provides a powerful tool for researchers and engineers to select the most informative sources to transfer knowledge. Except high-dimensional time-series modeling, our approach could also find applications in both sequential decision making and change point detection. For instance, transfer learning has arisen to handle the critical challenge of sparse feedbacks in reinforcement learning (RL), a popular framework for solving sequential decision making problems. However, negative transfer is still a notable challenge for multi-task reinforcement learning and it is risky to naively share information across all tasks \citep{TransferRL2023, MultiTaskRL2020}. Therefore, we embedded  our model with the offline RL transfer framework to automatically select correlated sources to share knowledge in a non-stationary environment \citep{NonStationaryRL2020}. In the future, we can further develop our method to account for an online RL task. Besides that, the estimated dynamic correlation among outputs can help us to understand the structure of time-series even with some missing data, as well as to detect some structural change points for subsequent decision making.}

Many extensions are possible for our model. Although the proposed methodology is flexible enough to capture complex and dynamic correlation, scaling it up to large datasets is computationally challenging. Possible solutions include utilizing a sparse approximation for the covariance matrix or developing a more efficient optimizing algorithm. 
In addition, the proposed EM optimization algorithm solely provides estimated values of the model parameters without incorporating any uncertainty measurement. To address this issue, we can use variational inference to approximate the true posterior distribution of parameters, thereby capturing the inherent uncertainty associated with these parameters. The approximated uncertainty would be propagated into the prediction at new points.


\newpage
\begin{APPENDICES}
\allowdisplaybreaks[4]

\section{DMGP-SS with missing data}
In general, observation may be missing at some time points within the range of $1 \leq t \leq n$. Under such a circumstance, the observation number for each output is $n_i \leq n$. To account for the missing data, we re-denote the data for the $i$th output as $\bm{X}_i = (\bm{x}_{i, t_1}, ..., \bm{x}_{i, t_i}, ..., \bm{x}_{i, t_{n_i}})^T$ and $\bm{y}_{i} = (y_{i, t_1}, ..., y_{i, t_i}, ..,y_{i, t_{n_i}})^T$, where $t_i$ represents the observation time index.

Compared with the model presented in the main text, the general DMGP-SS requires only adjustments on the parameter priors. The general model can be expressed as follows,  
\begin{align}
& \bm{y}_{(s)}, \bm{y}_{m} | \bm{\Phi}_{(s)},  \bm{\Phi}_{m} \sim  \mathcal{N} (\bm{0}, \bm{K}) \notag\\
& \alpha_{ii, t_i}| \alpha_{ii, t_{i-1}}  \sim {p}_{slab}(\alpha_{ii, t_{i}}| \alpha_{ii, t_{i-1}}),  \notag\\
&\bm{\theta}_{ii, t_{i}}| \bm{\theta}_{ii, t_{i-1}} \sim {p}_{slab}(\bm{\theta}_{ii, t_{i}}| \bm{\theta}_{ii, t_{i-1}}),  \notag\\
& \alpha_{im, t_{i}}|\gamma_{i, t_{i}}, \alpha_{im, t_{i-1}} \sim (1-\gamma_{i, t_{i}}) {p}_{spike}(\alpha_{im, t_{i}})  \notag\\
&\qquad \qquad \qquad \qquad +\gamma_{i, t_{i}} {p}_{slab}(\alpha_{im, t_{i}}| \alpha_{im, t_{i-1}}), \notag\\
& {\gamma}_{i, t_{i}}|{\eta}  \sim  Bern(\eta), \notag\\
&\bm{\theta}_{im, t_{i}}| \bm{\theta}_{im, t_{i-1}} \sim {p}_{slab}(\bm{\theta}_{im, t_{i}}| \bm{\theta}_{im, t_{i-1}}).
\label{eq: dynamic generative model with missing}
\end{align}
The spike prior is the same as that in the main text. The slab priors are re-defined as,
\begin{align}
{p}_{slab}^{hard}(\alpha_{im, t_i}| \alpha_{im, t_{i-1}}) = \frac{1}{2\nu_1} \exp \left(-\frac{| \alpha_{im, t_{i}} -  \alpha_{im, t_{i-1}}|}{\nu_1}\right),
\label{eq: hard slab with missing}
\end{align}
\begin{align}
{p}_{slab}^{soft}(\alpha_{im, t_{i}}| \alpha_{im, t_{i-1}}) =  \frac{1-\rho}{(1-\rho^{t_{i} - t_{i-1}})\sqrt{2 \pi \nu_1}} \exp \left(-\frac{( \alpha_{im, t_{i}} - \rho^{t_{i} - t_{i-1}} \alpha_{im, t_{i-1}})^2 (1-\rho^2)}{2\nu_1 (1-\rho^{2(t_{i} - t_{i-1})})}\right),
\label{eq: soft slab with missing}
\end{align}
where the soft prior is derived based on the property of an auto-regressive process.

\section{Proof of Proposition 1}
The proposed non-stationary MGP covariance matrix Eq. (9) is positive-definite, i.e., $\forall \bm{y} \neq \bm{0}$,
$$\bm{y}^T \bm{K} \bm{y} > 0.$$

\proof{Proof of Proposition 1}
Recall that the covariance functions are generated by the convolution of kernel functions:
\begin{align*}
{\rm cov}_{ii}^f(\bm{x}_t, \bm{x}_{t^{\prime}}) & = \alpha_{ii,t}\alpha_{ii,t'} \int g_{ii,t}(\bm{x}_{t} - \bm{u}) g_{ii, t'}(\bm{x}_{t'} - \bm{u}) d\bm{u} \notag\\
{\rm cov}_{im}^f(\bm{x}_t, \bm{x}_{t^{\prime}}) & = \alpha_{ii,t}\alpha_{im,t'} \int g_{ii,t}(\bm{x}_{t} - \bm{u}) g_{im, t'}(\bm{x}_{t'} - \bm{u}) d\bm{u} \notag\\
{\rm cov}_{mm}^f(\bm{x}_t, \bm{x}_{t^{\prime}}) & = \sum_{j=1}^m \alpha_{jm,t}\alpha_{jm,t'} \int g_{jm,t}(\bm{x}_{t} - \bm{u}) g_{jm, t'}(\bm{x}_{t'} - \bm{u}) d\bm{u}
\end{align*}

Decompose this quadratic form as follows,
\begin{align}
\bm{y}^T \bm{K} \bm{y}& = \sum_{1\leq i \leq m} \sum_{1\leq j \leq m} \sum_t \sum_{t'} y_{i,t}y_{j,t'} [{\rm cov}_{i,j}^f (\bm{x}_t, \bm{x}_{t^{\prime}}) + \sigma_i^2 \mathbb{I}_{i=j, t = t'}] \notag\\
&=  \sum_{1\leq i \leq m-1} \sum_t \sum_{t'} y_{i,t}y_{i,t'} {\rm cov}_{ii}^f (\bm{x}_t, \bm{x}_{t^{\prime}})  + 2 \sum_{1\leq i \leq m-1} \sum_t \sum_{t'} y_{i,t} y_{m, t'} {\rm cov}_{im}^f (\bm{x}_t, \bm{x}_{t^{\prime}}) \notag\\
& \quad + \sum_{t} \sum_{t'} y_{m,t} y_{m, t'} {\rm cov}_{mm}^f (\bm{x}_t, \bm{x}_{t^{\prime}}) + \sum_i \sum_t y_{i,t}^2 \sigma_i^2 \notag\\
&=  \sum_{1 \leq i \leq m-1} \left\{ \sum_t \sum_{t'} y_{i,t}y_{i,t'} \alpha_{ii,t}\alpha_{ii,t'} \int g_{ii,t}(\bm{x}_{t} - \bm{u}) g_{ii, t'}(\bm{x}_{t^{\prime}} - \bm{u}) d\bm{u} \right. \notag\\
& \qquad \qquad \qquad + \sum_t \sum_{t'} y_{m,t}y_{m,t'} \alpha_{im,t}\alpha_{im,t'} \int g_{im,t}(\bm{x}_{t} - \bm{u}) g_{im, t'}(\bm{x}_{t^{\prime}} - \bm{u}) d\bm{u} \notag\\
& \qquad \qquad \qquad \left. + 2\sum_t \sum_{t'} y_{i,t}y_{m,t'} \alpha_{ii,t}\alpha_{im,t'} \int g_{ii,t}(\bm{x}_{t} - \bm{u}) g_{im, t'}(\bm{x}_{t^{\prime}} - \bm{u}) d\bm{u} \right\} \notag\\
& \quad + \sum_t \sum_{t'} y_{m,t}y_{m,t'} \alpha_{mm,t}\alpha_{mm,t'} \int g_{mm,t}(\bm{x}_{t} - \bm{u}) g_{mm, t'}(\bm{x}_{t^{\prime}} - \bm{u}) d\bm{u} + \sum_i \sum_t y_{i,t}^2 \sigma_i^2 \notag\\
&=  \sum_{1 \leq i \leq m-1} \left\{  \int \left[ \sum_t  y_{i,t}\alpha_{ii,t} g_{ii,t}(\bm{x}_{t} - \bm{u}) \sum_{t'} y_{i,t'} \alpha_{ii,t'} g_{ii, t'}(\bm{x}_{t^{\prime}} - \bm{u})  \right. \right. \notag\\
& \qquad \qquad \qquad \qquad +  \sum_t y_{m,t} \alpha_{im,t} g_{im,t}(\bm{x}_{t} - \bm{u}) \sum_{t'} y_{m,t'}\alpha_{im,t'}  g_{im, t'}(\bm{x}_{t^{\prime}} - \bm{u})  \notag\\
& \qquad \qquad \qquad \qquad \left. \left. + 2 \sum_t y_{i,t}  \alpha_{ii,t} g_{ii,t}(\bm{x}_{t} - \bm{u}) \sum_{t'} y_{m,t'} \alpha_{im,t'} g_{im, t'}(\bm{x}_{t^{\prime}} - \bm{u}) \right] d\bm{u} \right\} \notag\\
& \quad + \int  \sum_t  y_{m,t}  \alpha_{mm,t} g_{mm,t}(\bm{x}_{t} - \bm{u}) \sum_{t'}y_{m,t'}\alpha_{mm,t'} g_{mm, t'}(\bm{x}_{t^{\prime}} - \bm{u}) d\bm{u} + \sum_i \sum_t y_{i,t}^2 \sigma_i^2 \notag\\
&=  \sum_{1 \leq i \leq m-1} \left\{  \int \left[ \sum_t  y_{i,t}\alpha_{ii,t} g_{ii,t}(\bm{x}_{t} - \bm{u}) + \sum_{t'} y_{m,t'} \alpha_{im,t'} g_{im, t'}(\bm{x}_{t^{\prime}} - \bm{u}) \right]^2 d\bm{u} \right\} \notag\\
& \quad + \int  \left[ \sum_t  y_{m,t}  \alpha_{mm,t} g_{mm,t}(\bm{x}_{t} - \bm{u}) \right]^2 d\bm{u} + \sum_i \sum_t y_{i,t}^2 \sigma_i^2  > 0 
\label{eq: quadratic decompostion 1}
\end{align}
Proof completes.
\endproof

\section{Derivation of the objective function in M-step}
Based on Bayes theorem, the parameter posterior can be expressed as:
\begin{align*}
p({\bm{\Phi}}, \bm{\gamma} | \bm{y}) \propto p(\bm{y}|\bm{\Phi})p(\bm{\Phi}|\bm{\gamma})p(\bm{\gamma}).
\end{align*}
And based on the theory of multivariate Gaussian distribution, we have:
\begin{align}
&p(\bm{y}|\bm{\Phi}) = p(\bm{y}_{(s)}, \bm{y}_m | \bm{\Phi}_{(s)}, \bm{\Phi}_m) \notag\\
&=\mathcal{N}
\left(
\begin{bmatrix} \begin{array} {c}
\bm{y}_{(s)} \\ \bm{y}_m
\end{array} \end{bmatrix}\Big |
\begin{bmatrix} \begin{array} {c}
\bm{0} \\ \bm{0}
\end{array} \end{bmatrix},
\begin{bmatrix} \begin{array}{cc}
\bm{K}_{(ss)} & \bm{K}_{(sm)} \\ \bm{K}_{(sm)}^T & \bm{K}_{ mm}
\end{array} \end{bmatrix}
\right) \notag\\
&= \mathcal{N} \left( \bm{y}_{(s)} | \bm{0}, \bm{K}_{(ss)}  \right) \mathcal{N} \left( \bm{y}_{m} | \bm{\mu}, \bm{\Sigma}  \right),
\end{align}
where $\bm{\mu}=\bm{K}_{(sm)}^T \bm{K}_{(ss)}^{-1}\bm{{y}}_{(s)}$ is the conditional mean of target given the sources and $\bm{\Sigma}=\bm{K}_{mm}-\bm{K}_{(sm)}^T \bm{K}_{(ss)}^{-1}\bm{K}_{(sm)}$ is the conditional covariance. 

Therefore, the objection function can be derived as:
\begin{align}
&E_{\bm{\gamma}} \left\{ \log p({\bm{\Phi}}, \bm{\gamma} | \bm{y}) \right\} \notag\\
=& E_{\bm{\gamma}} \left\{ \log p(\bm{y} | {\bm{\Phi}}, \bm{\gamma}) p({\bm{\Phi}}, \bm{\gamma}) \right\}  + const. \notag\\
=& \log p(\bm{y}_{(s)}|  \bm{\Phi}_{(s)} )  +  \log  p(\bm{y}_m | \bm{\Phi}_m, \bm{y}_{(s)}, {\bm{\Phi}}_{(s)})   \notag\\
&+ \log p(\bm{\Phi}_{(s)}) + E_{\bm{\gamma}} \left\{ \log p(\bm{\Phi}_m| \bm{\gamma}) + \log p(\bm{\gamma}) \right\}  + const. \notag\\
=& \log p(\bm{y}_{(s)}|  \bm{\Phi}_{(s)} )  +  \log  p(\bm{y}_m | \bm{\Phi}_m, \bm{y}_{(s)}, {\bm{\Phi}}_{(s)})   \notag\\
&+ \log p(\bm{\theta}_{(s)})  + \log p(\bm{\alpha}_{(s)})  + \log p(\bm{\theta}_{m}) + E_{\bm{\gamma}} \left\{ \log p(\bm{\alpha}_m| \bm{\gamma}) \right\}  + const. \notag\\
=& -\frac{1}{2} \left\{ \bm{y}_{(s)}^T \bm{K}_{(ss)}^{-1} \bm{y}_{(s)} + \log|\bm{K}_{(ss)}| + (\bm{y}_{m}-\bm{\mu})^T \bm{\Sigma}^{-1} (\bm{y}_{m}-\bm{\mu}) + \log|\bm{\Sigma}| \right\} \notag\\
& + \sum_{i=1}^{m-1} \sum_{t=2}^n \Big[  \log {p}_{slab}(\bm{\theta}_{ii, t}| \bm{\theta}_{ii, t-1}) + \log {p}_{slab}({\alpha}_{ii, t}| {\alpha}_{ii, t-1}) + \log {p}_{slab}(\bm{\theta}_{im, t}| \bm{\theta}_{im, t-1})  \Big] \notag\\
& + \sum_{i=1}^m \sum_{t=2}^n \Big[  (1-E_{\bm{\gamma}}{\gamma}_{i, t}) \log {p}_{spike}(\alpha_{im, t}) + E_{\bm{\gamma}}{\gamma}_{i, t} \log {p}_{slab}(\alpha_{im, t}| \alpha_{im, t-1}) \Big] + const.. 
\label{eq: EM objective derivation}
\end{align}

\section{Details of DMGP-GP}
\textbf{DMGP-GP} is a state-of-art non-stationary MGP model, which constructs a LMC model for all outputs and assumes the hyper-parameters follow other GPs \cite{Non-stationaryMGP2021}:
\begin{align}
\bm{y}({x}_t) &= \bm{A}_t \bm{q}({x}_t) + \bm{\epsilon} \notag\\
\log(A_{ii, t}) &\sim \mathcal{GP}(0, k_{\alpha}(t, t^{\prime})) \notag\\
A_{ij, t} &\sim \mathcal{GP}(0, k_{\alpha}(t, t^{\prime})), i \neq j \notag\\
q_i ({x}_t) &\sim \mathcal{GP}(0, k({x}_t, {x}_{t^{\prime}})),  \notag\\
k(x_t, x_{t^{\prime}}) &= \sqrt{\frac{2 \theta_{t} \theta_{t^{\prime}}} {\theta_{t}^2 + \theta_{t{\prime}}^2}} \exp \left[ \frac{(x_t - x_{t^{\prime}})^2}{2(\theta_{t}^2 + \theta_{t^{\prime}}^2)} \right] \notag\\
\log(\theta_{t}) & \sim \mathcal{GP}(0, k_{\theta}(t, t^{\prime}))
\end{align}
where $\bm{y}(x_t)  = [y_1(x_t), ..., y_m(x_t)]^T$ are $m$ outputs, $\bm{A}_t \in R^{m \times m}$ is the time-varying coefficient matrix, $\bm{q}(x_t) = [q_1(x_t), ..., q_m(x_t)]^T$ are $m$ i.i.d. latent Gaussian processes with zero mean and the same covariance function $k(x_t, x_{t^{\prime}})$, and $\bm{\epsilon}=(\epsilon_1,...\epsilon_m)$ is measurement noise with $\epsilon_i \sim N(0, \sigma_i^2)$. The covariance for the $m$ outputs is
$${\rm cov}[\bm{y}(x_t),  \bm{y}(x_{t^{\prime}})] = \bm{A}_t \bm{A}_{t'}^T k(x_t, x_{t^{\prime}}) + {\rm diag}\{\sigma_i\},$$
where $\bm{A}_t \bm{A}_{t^{\prime}}^T \in \mathbb{R}^{m \times m}$ is the correlation matrix of $m$ outputs, and ${\rm diag}\{\sigma_i\}$ is the diagonal matrix with elements $\{\sigma_i\}_{i=1}^m$.

\section{Scalability of DMGP-SS}
Since our model supposes that the latent processes $z_i(\bm{x})$ are independent on each other, the computational complexity is $O(mn^3)$ when all the $m$ outputs have an equal length of $n$. In comparison, the computational complexity of the classical MGP is $O(m^3n^3)$, much larger than that of ours. Therefore, the proposed model can handle a number of outputs much easier. 
For example, we test our method on one numerical case with up to $mn=8580$ points. \Tabref{tab: scalable experiments} show the prediction error and model fitting time. The prediction accuracy of the proposed method is better than that of GP in all the three experiments. Besides, although the third experiments have two times as many points as the first one, the fitting time of the third one ($m=33, n=260$) is only two times that of the first one ($m=17, n=260$), which identifies that the computational complexity of our method is $O(mn^3)$. Besides, the fitting time of the second experiment ($m=17, n=520$) is only four times that of the first one ($m=17, n=260$), which means the second experiment takes less gradient descent steps to converge than the first one does. 
\begin{table}[H]
    \centering
    \caption{Prediction error and fitting time of DMGP-SS with up to 8580 points.}
    \begin{tabular}{c c ccc}
        \hline
        & & m=17, n=260 & m=17, n=520 & m=33, n=260  \\
        \hline
        GP & MAE & 0.803 & 0.903 & 0.781 \\
        \multirow{2}{*}{DMGP-SS} & MAE & 0.601 & 0.432 & 0.606 \\
        & Time (seconds) & (508) & (1938) & (1273) \\
        \hline
    \end{tabular}
    \label{tab: scalable experiments}
\end{table}
\end{APPENDICES}

\ACKNOWLEDGMENT{This work was supported by NSFC under Grants NSFC-72171003, NSFC-71932006.}





\end{document}